\documentclass[twoside]{article}

\usepackage{amazon_paper}

\usepackage{amssymb}
\usepackage{xcolor}
\usepackage{booktabs}
\usepackage[]{graphicx}
\usepackage{caption}
\usepackage{subcaption}
\usepackage{wrapfig}
\definecolor{amazondarkblue}{HTML}{232F3E}

\usepackage{multirow}

\usepackage[table]{xcolor}
\usepackage{multirow}
\usepackage{array}
\definecolor{baselinegray}{gray}{0.92}
\definecolor{overallbg}{RGB}{238,242,255}
\newcolumntype{O}{>{\columncolor{baselinegray}}c}

\usepackage[ruled,vlined,linesnumbered]{algorithm2e}
\SetKwComment{tcc}{\color{gray}\# }{}
\SetKwComment{tcp}{\color{gray}// }{}
\DontPrintSemicolon

\usepackage{pifont}
\newcommand{\cmark}{\ding{51}} 
\newcommand{\xmark}{\ding{55}} 
\usepackage{makecell} 
\usepackage[dvipsnames]{xcolor} 
\newcommand{\gain}[1]{\textcolor{ForestGreen}{#1}}
\newcommand{\loss}[1]{\textcolor{BrickRed}{#1}}

\usepackage{listings}
\usepackage{xcolor}

\lstdefinelanguage{json}{
    basicstyle=\ttfamily\small,
    comment=[l]{//},
    morestring=[b]",
    stringstyle=\color{blue},
    literate=
     *{0}{{{\color{black}0}}}{1}
      {1}{{{\color{black}1}}}{1}
      {2}{{{\color{black}2}}}{1}
      {3}{{{\color{black}3}}}{1}
      {4}{{{\color{black}4}}}{1}
      {5}{{{\color{black}5}}}{1}
      {6}{{{\color{black}6}}}{1}
      {7}{{{\color{black}7}}}{1}
      {8}{{{\color{black}8}}}{1}
      {9}{{{\color{black}9}}}{1}
}

\usepackage[most]{tcolorbox}
\usepackage{courier} 

\newtcolorbox{promptbox}[2][]{%
  enhanced,
  breakable,
  colback=gray!2,
  colframe=gray!30,
  boxrule=0.4pt,
  arc=6pt,
  outer arc=2pt,
  fontupper=\ttfamily\small,
  title={#2},
  coltitle=white,
  colbacktitle=gray!80!black,
  attach boxed title to top left={yshift=-1.5mm, xshift=2mm},
  boxed title style={
    sharp corners,
    boxrule=0pt,
    colback=gray!80!black,
    colupper=white,
    fontupper=\bfseries\footnotesize,
  },
  #1
}

\begin{document}

\title{%
\raisebox{-2.5em}{%
  \parbox[t]{1.0in}{\includegraphics[width=0.9in]{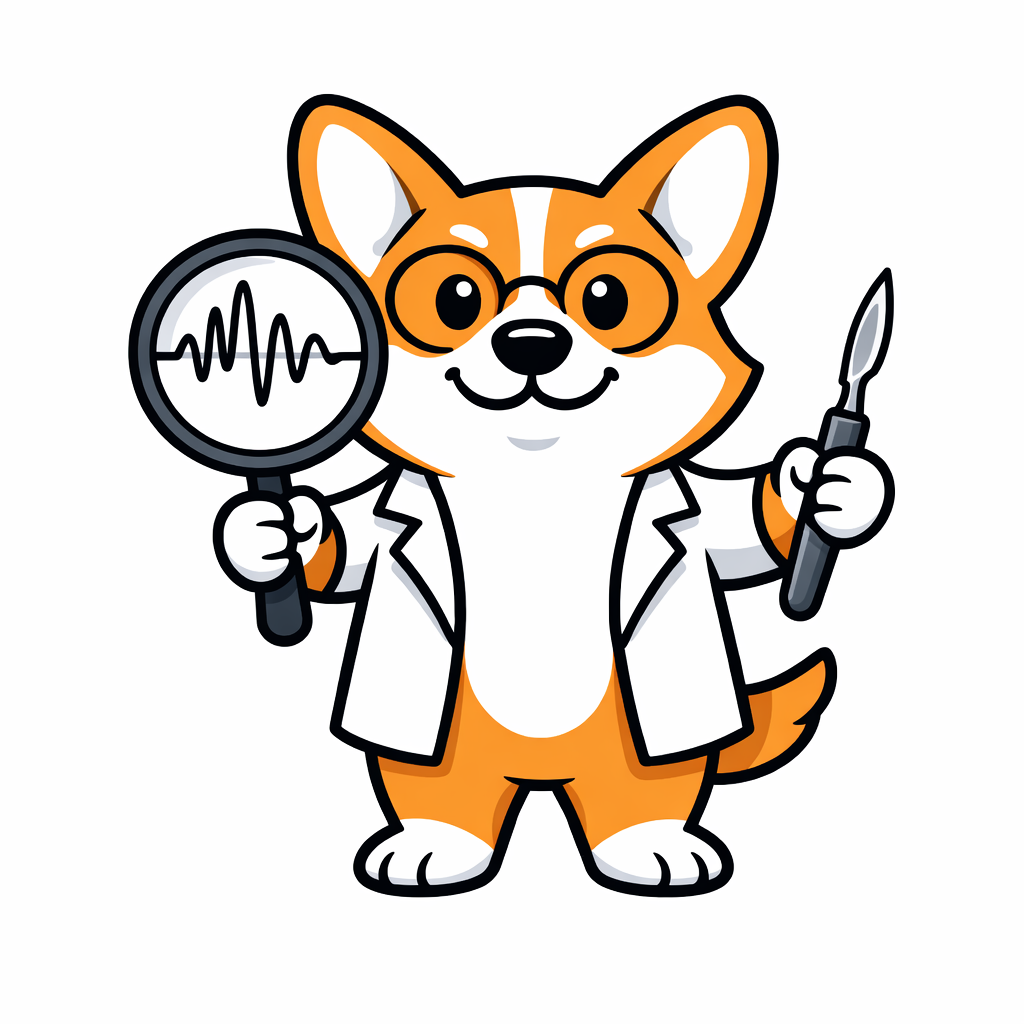}}%
}%
\hspace{0.1in}%
\parbox[t]{0.7\textwidth}{\centering SenTSR-Bench: Thinking with Injected Knowledge for Time-Series Reasoning}%
}

\author{%
\small{Zelin He$^{1}$\thanks{Work done during an internship at Amazon.}, Boran Han$^{2}$, Xiyuan Zhang$^{2}$, Shuai Zhang$^{2}$, Haotian Lin$^{3}$,\\
Qi Zhu$^{2}$, Haoyang Fang$^{2}$, Danielle C.\ Maddix$^{2}$, Abdul Fatir Ansari$^{2}$,\\
Akash Chandrayan$^{3}$, Abhinav Pradhan$^{3}$, Bernie Wang$^{2}$, Matthew Reimherr$^{1,3}$}%
  \\[1em]
  {\fontsize{10pt}{11pt}\selectfont
$^{1}$The Pennsylvania State University \quad
$^{2}$AWS AI Labs \quad
$^{3}$Amazon RME}\\[0.3em]
  {\fontsize{10pt}{11pt}\selectfont \href{https://zlhe0.github.io/SenTSR-Bench-Website/}{\raisebox{-0.1em}{\includegraphics[height=1em]{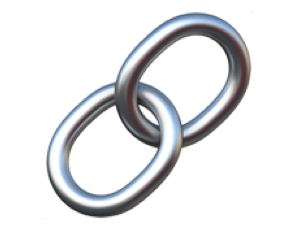}}\,Project Page}}
}

\maketitle
\vspace{-1.5em}
\definecolor{amazonorange}{HTML}{FF9900}
\renewenvironment{abstract}{%
  \begin{tcolorbox}[colback=amazonorange!8, colframe=black, boxrule=0.5pt, arc=3pt, left=10pt, right=10pt, top=8pt, bottom=8pt]
  \centerline{\textbf{Abstract}}\vspace{0.5em}
}{%
  \end{tcolorbox}
}
\begin{abstract}
Time‐series diagnostic reasoning is essential for many applications, yet existing solutions face a persistent gap: general reasoning large language models (GRLMs) possess strong reasoning skills but lack the domain-specific knowledge to understand complex time-series patterns. Conversely, fine-tuned time-series LLMs (TSLMs) understand these patterns but lack the capacity to generalize reasoning for more complicated questions. To bridge this gap, we propose a hybrid \emph{knowledge-injection} framework that injects TSLM-generated insights directly into GRLM's reasoning trace, thereby achieving strong time-series reasoning with in-domain knowledge. As collecting data for knowledge injection fine-tuning is costly, we further leverage a reinforcement learning-based approach with verifiable rewards (RLVR) to elicit knowledge-rich traces \textit{without human supervision}, then transfer such an in-domain thinking trace into GRLM for efficient knowledge injection. We further release \emph{SenTSR-Bench}, a {multivariate} time-series-based diagnostic reasoning benchmark collected from \textit{real-world industrial operations}. Across \emph{SenTSR-Bench} and other public datasets, our method consistently surpasses TSLMs by {9.1\%–26.1\%} and GRLMs by {7.9\%–22.4\%}, delivering robust, context-aware time-series diagnostic insights.
\end{abstract}

\section{Introduction}

\begin{figure*}[htbp]
    \centering
    \includegraphics[width=1.0\linewidth]{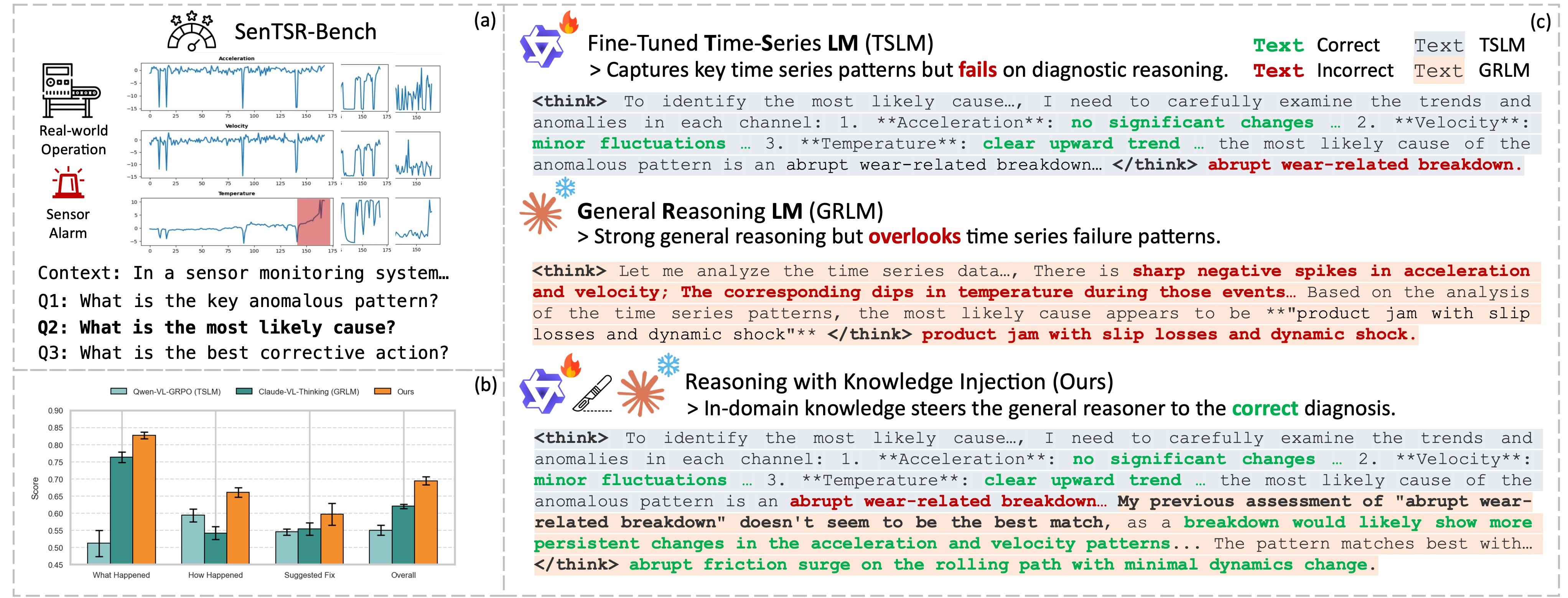}
    \caption{(a) The newly released \emph{SenTSR-Bench} benchmark, collected from real-world machine monitoring environments, with multi-stage diagnostic questions. 
(b) Performance of the proposed framework on \emph{SenTSR-Bench}, surpassing both stand-alone time-series specialists (TSLM) and general reasoning models (GRLM). 
(c) Case study illustrating why knowledge injection helps: the \emph{specialist} captures key time-series patterns but fails to connect them to the correct root cause; the \emph{general reasoner} shows strong reasoning but overlooks domain-specific critical failure patterns; our method injects the \textit{in-domain knowledge} from fine-tuned specialist into the reasoner’s \textit{reasoning trace}, aligning the trace with domain knowledge and producing the correct diagnosis.}
    \label{fig:case_study}
\end{figure*}

Diagnostic reasoning over time-series data is a fundamental capability in many domains, enabling critical tasks such as event characterization, root-cause diagnosis, and decision-making \citep{leite2024fault,chen2024artificial}. In industrial operations, for instance, streams of sensor data measuring machine temperature and vibration are analyzed to diagnose potential equipment failures (Figure~\ref{fig:case_study} (a)). However, existing research in this domain has predominantly focused on surface-level anomaly detection \citep{alnegheimish2025m}.  While effective at identifying irregularities, these techniques cannot offer actionable insights because they lack the capacity for temporal and causal reasoning required to explain an anomaly's origin, diagnose its root cause, or recommend corrective actions.

Recent advances in LLMs \citep{jaech2024openai,guo2025deepseek,anthropic2025claude37sonnet}, have unlocked an enhanced reasoning capabilities via embed implicit reasoning mechanisms \citep{yeo2025demystifying}, yielding remarkable gains on benchmarks requiring reasoning. However, these general reasoning LLMs (GRLMs) lack the domain knowledge needed to interpret complex time-series patterns, thereby producing incorrect reasoning trajectories and thus incorrect diagnoses \citep{merrill2024language,cao2026more}. In parallel, smaller LLM variants fine‐tuned on domain‐specific time‐series-textual pairs \citep{xie2024chatts, zhang2025timemaster} have shown improved alignment with time-series understanding tasks. Yet these fine-tuned time-series language models (TSLMs) frequently overfit to narrow template, like tasks and lack the reasoning depth or generalization capacity required for out-of-distribution scenarios. As a result, \textit{both standalone GRLMs and TSLMs fall short in practice} (illustrated in Figure~\ref{fig:case_study}(c)).

To address the above challenge, we propose a \emph{reasoning with knowledge injection} framework that couples the reasoning power of GRLMs with the in-domain knowledge of TSLMs. At its core, the framework injects knowledge from TSLMs directly into the reasoning process of GRLMs, allowing the generated reasoning trace to continue with guidance from in-domain information. When the injected knowledge is reliable, it helps steer the reasoning trajectory toward accurate diagnoses; when the knowledge is weaker, the model corrects it with its strong critical thinking capacity. 

One additional challenge is that a TSLM trained for in-domain question answering often fails to function effectively as an assistant for a GRLM. A typical alternative is to finetune a dedicated helper model, but this approach is constrained by the need to construct large, high-quality datasets explicitly tailored for knowledge injection. To overcome this supervision bottleneck, we introduce thinking transfer. Our method trains the TSLM within a  reinforcement learning with verifiable reward framework \citep{guo2025deepseek}, leveraging rule-based verifiable rewards and an explicit thinking structure to naturally elicit knowledge-rich thinking traces without any manual supervision. At inference, these RL-honed traces are injected into the GRLM, providing it with high-quality, in-domain knowledge to ground its subsequent reasoning process. 

Furthermore, to benchmark time‐series diagnostic reasoning in real-world diagnostic settings, we introduce \underline{Sensor}-based \underline{T}ime-\underline{S}eries Diagnostic \underline{R}easoning (\emph{SenTSR-Bench}) Benchmark, a first-of-its-kind dataset of multivariate sensor streams and diagnostic texts for time-series diagnostic reasoning evaluation. In contrast to prior benchmarks that are either purely synthetic or LLM-annotated, \emph{SenTSR-Bench} is built on the \textit{real-world multivariate time-series data} drawn from real-world diagnostics events with \textit{human-annotated data}.

Across \textit{SenTSR-Bench} and other existing benchmark datasets, and on both closed-source and open-source reasoning models, our method surpasses TSLMs by 9.1–26.1\% and GRLMs by 7.9–22.4\%. RL-enhanced injection further yields 1.66×–2.92× larger gains than SFT-enhanced injection, and consistently outperforms few-shot prompting and prompt-based collaboration approaches. Taken together, our key contributions are as follows:

\begin{figure*}[!t]
    \centering
    \includegraphics[width=\linewidth]{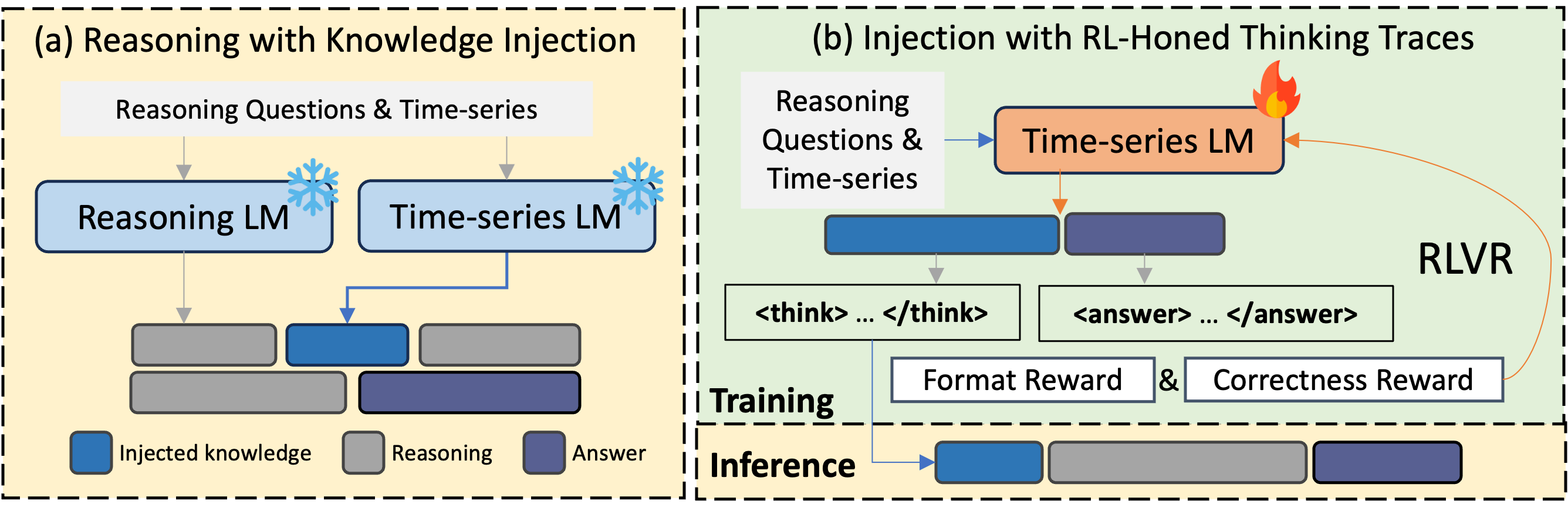}
    \caption{{Overview of the proposed paradigm.}
(a) \textit{ Knowledge injection:} given a reasoning question and its time-series, a time–series LM (TSLM) produces grounded analysis snippets that are injected into the reasoning trace of a general \emph{frozen} reasoning LM (GRLM) to answer diagnostic queries without weight updates.
(b) \textit{Thinking transfer via RL:}  We train the TSLM using reinforcement learning with \emph{verifiable rewards} (RLVR) with an explicit thinking structure to \emph{elicit} analysis-first thinking traces \emph{without human supervision}; at inference, these traces are transferred via injection into the reasoning LM to strengthen temporal grounding for diagnosis.}
    \label{fig:workflow}
\end{figure*}

$\bullet$ \textbf{New Paradigm for Time-series Reasoning.} We formalize a framework that injects in-domain knowledge from a TSLM into an GRLM’s reasoning process, steering reasoning with domain knowledge.\\
$\bullet$ \textbf{RL-Based Method for Efficient Injection.} We propose an injection paradigm that utilizes reinforcement learning with verifiable rewards to elicit knowledge‐rich thinking traces \emph{without manual supervision} for injection.\\
$\bullet$ \textbf{Real-World Benchmark and Evaluation.} We release \emph{SenTSR-Bench}, a de-identified, real-world multivariate time-series benchmark for diagnostic reasoning. Evaluations on \emph{SenTSR-Bench} and public datasets show state-of-the-art diagnostic accuracy of our proposed solution with interpretable explanations.

\section{Methodology}
Figure~\ref{fig:workflow} provides an overview of our proposed framework. In this section, we first establish preliminaries and formally define the reasoning model generation process (Section~\ref{method:prelim}). We then introduce the general paradigm of knowledge injection (Section~\ref{method:paradigm}). We then further instantiate this framework (Section~\ref{method:early}). Finally, we describe a reinforcement learning-based framework for efficient knowledge injection (Section~\ref{method:rl}).

\subsection{Preliminaries and Notation}
\label{method:prelim}

\paragraph{Multimodal Input.}  
Write \(V\) for the discrete token vocabulary and \(V^{*}\) for the space of finite token sequences, and use \([a,b]\) to denote the concatenation of two sequences \(a\) and \(b\). Let \(\mathbf{q}=(q_1,\ldots,q_{n}) \in V^{*}\) be a sequence of textual tokens describing the task (e.g., question, context, or instructions).   
A multivariate time-series is denoted by \(\mathbf{X}=\{\mathbf{x}_t\}_{t=1}^{T}\), where each \(\mathbf{x}_t \in \mathbb{R}^{D}\) is the reading of \(D\) channels at time step \(t\). To interface with language models, \(\mathbf{X}\) must be mapped into the token space \(V^{*}\). This can be done, for example, by rendering the series as a line-plot image and encoding it \citep{liu2025picture}, converting it into structured JSON text followed by standard text tokenization, or applying a specialized time-series tokenizer \citep{xie2024chatts}. With slight abuse of notation, we use \(\mathbf{X}\) to denote the final tokenized representation.

\paragraph{Reasoning Model.} 
We define a reasoning model through its generative distribution \(\pi\) (also referred to as a policy in later context) that generates two outputs: an internal reasoning trace \(\mathbf{r}=(r_1,\ldots,r_K)\in V^{*}\) and a final answer \(\mathbf{y}=(y_1,\ldots,y_M)\in V^{*}\). Generation proceeds in two phases. In the reasoning phase, the model autoregressively produces a latent reasoning trace conditioned on the input pair \((\mathbf{X},\mathbf{q})\) and a special thinking structure:
\begin{align}
\pi(\mathbf{r}\mid \mathbf{X},\mathbf{q})
= \prod_{k=1}^{K}
\pi\!\left(r_k \mid  \mathbf{X}, \mathbf{q}, [\langle\mathrm{think}\rangle, \mathbf{r}_{<k}]\right).
\end{align}
Here, \(\langle\mathrm{think}\rangle\) marks the beginning of the reasoning segment, which continues until the model emits the closing token \(\langle/\mathrm{think}\rangle\). In the response phase, the model conditions on both the input and the full reasoning trace to generate the final answer:
\begin{align}
\label{eq:answer_generation}
\pi(\mathbf{y}\mid \mathbf{X},\mathbf{q},[\langle\mathrm{think}\rangle, \mathbf{r},\langle\mathrm{/think}\rangle]) = \prod_{j=1}^{M} \pi\Big(
y_j \,\Big|\, \mathbf{X}, \mathbf{q}, [\langle\mathrm{think}\rangle,\mathbf{r}, \langle/\mathrm{think}\rangle,\mathbf{y}_{<j}]\Big).
\end{align}

This reasoning–then–response decomposition exposes the latent reasoning trace \(\mathbf{r}\), which we later \emph{inspect} and \emph{modify} through knowledge injection.

In this paper, we distinguish two models. A (frozen) general reasoning model (GRLM), quantified by \(\pi^{G}\), is a large open/closed–source model that follows the reasoning–then–response factorization discussed above, and a time-series language model (TSLM), quantified by \(\pi^{T}\), is a small fine–tuned in-domain specialist. 

\subsection{General Knowledge Injection Paradigm}
\label{method:paradigm}

\paragraph{Specialist Knowledge Generation.} Given the current reasoning state of the general reasoner \(\pi^{G}\) at step \(k\), i.e., the prefix \(\mathbf{r}^{G}_{<k}\) together with inputs \((\mathbf{X},\mathbf{q})\), we form an \emph{injection–oriented token sequence}
\(
\mathbf{\tilde{q}} \;=\; \mathsf{Query} \!\big(\mathbf{q},\, \mathbf{r}^{G}_{\le k}\big),
\)
where \(\mathsf{Query}(\cdot)\) is a deterministic query–shaping function (e.g., “provide helpful information”, “validate the claims”). Concrete choices are given in later subsections. Then a TSLM is invoked on \((\mathbf{X},\mathbf{\tilde{q}})\) to produce an output sequence
\[
\mathbf{K}^{T} \;\sim\; \pi^{T}\big(\,\cdot \,\big|\, \mathbf{X},\, \mathbf{\tilde{q}}\big).
\]
Intuitively, $\mathbf{K}^{T}$ stands for the relevant in-domain time-series knowledge for injection.

\paragraph{Reasoning with Knowledge Injection.}
Given the TSLM knowledge output \(\mathbf{K}^{T}\) and the current GRLM reasoning prefix \(\mathbf{r}^{G}_{\le k}\), we apply injection with 
\[
\mathbf{r}^{\text{Inj}}_{\le k}\;=\;\mathsf{Inject} \big(\,\mathbf{r}^{G}_{\le k},\;\mathbf{K}^{T}\, \big),
\]
which returns an updated thinking trace prefix to be used. Here \(\mathsf{Inject}(\cdot)\) is a deterministic injection function; the choice of \(k\) is chosen based on the injection method. The general reasoner GRLM then resumes generation conditioned on the updated prefix:
\begin{align}
\label{eq:GRLM}
r^{G}_{j} \;&\sim\; \pi^{G}\!\Big(\,\cdot \;\Big|\;  \mathbf{X},\; \mathbf{q},[\langle\mathrm{think}\rangle,\; \mathbf{r}^{\text{Inj}}_{<j}]\;\Big),
\qquad j \ge k,
\end{align}
and then produce the final answer similar to Eq. (\ref{eq:answer_generation}).

\subsection{Instantiating Knowledge Injection}
\label{method:early}

\paragraph{Early Knowledge Injection} A simple yet effective way to realize the injection paradigm is through early injection: immediately after \(\langle\mathrm{think}\rangle\).
We choose a {tokenized} instruction \(\mathbf{v}_{\mathrm{help}}\) (e.g., “produce a step–by–step analysis of the question with the time-series data”) and form
\[
\mathbf{\tilde{q}} \;=\; \mathsf{Query}_{\mathrm{help}}\!\big(\mathbf{q},\,\emptyset\big) = \left[\mathbf{q},\mathbf{v}_{\mathrm{help}}\right].
\]
The specialist then generates the knowledge snippet
\(
\mathbf{K}^{T} \sim \pi^{T}(\cdot \mid \mathbf{X},\mathbf{\tilde{q}})
\) from the learnt time-series knowledge,
which we then appended a brief reflection trigger to elicit critical reasoning $\mathbf{v}_{reflect}$ (e.g., “Wait, let me reflect on my previous thinking process with the time-series data.”). We then inject at \(k{=}1\):
\begin{align*}
\mathbf{r}^{\mathrm{Inj}}_{\le 1}
\;=\;
\mathsf{Inject}_{reflect} \big(\,\emptyset,\;{\mathbf{K}}^{T}\,\big)=
\big[\; {\mathbf{K}}^{T}, \boldsymbol{v}_{reflect}\,\big], 
\end{align*}
after which the general reasoner \(\pi^{G}\) continues its reasoning trace and produces the final response conditioned on \(\mathbf{r}^{\mathrm{Inj}}_{\le 1}\) (cf.\ Section \ref{method:prelim}). Conceptually, \(\pi^{T}\) contributes \emph{grounded, in–domain time-series-based insights} extracted from \(\mathbf{X}\), while \(\pi^{G}\) performs the \emph{general reasoning} by integrating the injected knowledge with context \(\mathbf{q}\),  adjudicating alternatives, and producing the final answer. 

\paragraph{Other Injection Paradigms}
Beyond early injection, the framework also supports alternative strategies. Examples include \textit{intermediate injection} that corrects the GRLM's reasoning process by inserting TSLM's knowledge at low-confidence points in the reasoning trace; or \textit{late injection} that prompts TSLM to critique the entire GRLM reasoning trace and prompts reflection before the final answer. Full implementation details are provided in Appendix~\ref{app:implementation}. In practice, we find that early injection is the most broadly effective, and thereby we adopt early injection as the default in subsequent method development, and report comparison results for the other variants in Section \ref{sec:framework_analysis}.

\paragraph{Practical Implementation.}
The method is easy to implement and compatible with standard LLM APIs. For models that support assistant prefill, the injected trace can be directly fed as assistant's initial tokens by pre-inserting
\(
[\,\langle\mathrm{think}\rangle,\;\mathbf{r}^{\text{Inj}}_{\le k}\,]
\). For models do not allow prefill for reasoning traces, we instead use an instructional proxy by wrapping the injected trace in the model’s recommended thinking templates. See Appendix~\ref{app:implementation} for details.

\begin{algorithm}[!t]
\caption{Algorithm Workflow for Knowledge Injection with RL Honed Thinking}
\label{alg:ki-rl}
\KwIn{Training set $\mathcal{D}_{\mathrm{train}}$, test set $\mathcal{D}_{\mathrm{test}}$, general reasoner policy $\pi^{G}$}
\KwOut{Trained specialist policy $\pi^{T}$ and predictions on $\mathcal{D}_{\mathrm{test}}$}

\BlankLine
\tcc{Stage I: Train TSLM with RLVR}
\For{$(\mathbf{X}, \mathbf{q}, \mathbf{y}^{\star}) \in \mathcal{D}_{\mathrm{train}}$}{
  Update $\pi^{T}$ using RLVR training with composite reward \tcp*{cf. Eq.~(\ref{eq:grpo_main})}
}

\BlankLine
\tcc{Stage II: Inference-time knowledge injection for GRLM}
\For{$(\mathbf{X}, \mathbf{q}) \in \mathcal{D}_{\mathrm{test}}$}{
  Obtain $\mathbf{r}^{T} \sim \pi^{T}(\cdot \mid \mathbf{X}, \mathbf{q})$\;
  Form $\mathbf{r}^{\mathrm{Inj}}_{\le 1} \leftarrow \mathsf{Inject}_{reflect}(\emptyset, \mathbf{r}^{T})$\tcp*{cf. Eq.~(\ref{eq:early_injection})}
  Obtain $\mathbf{r}^{G} \sim \pi^{G}(\cdot \mid \mathbf{X},\; \mathbf{q},[\langle\mathrm{think}\rangle,\mathbf{r}^{\mathrm{Inj}}_{\le 1}])$\;
  Produce and record $\mathbf{y}^{G}$ with $\mathbf{X}, \mathbf{q}, \mathbf{r}^{G}$\tcp*{cf. Eq.~(\ref{eq:answer_generation})}
}
\Return{$\pi^{T}$ and all test predictions}
\end{algorithm}

\subsection{Knowledge Injection with RL-Honed Thinking Traces}
\label{method:rl}

A time-series specialist \(\pi^{T}\) is typically optimized for direct question answering,  
\[
\mathbf{y}^{T} \sim \pi^{T}\!\left(\,\cdot \;\middle|\; \mathbf{X},\, \mathbf{q}\right),
\]  
where the objective is to predict the answer tokens \(\mathbf{y}^{T}\) given inputs \((\mathbf{X},\mathbf{q})\).  
In contrast, knowledge injection requires the specialist to provide an intermediate analysis or evidence rather than a final answer. This is usually elicited through a help-oriented query,  
\[
\mathbf{K}^{T} \sim \pi^{T}\!\left(\,\cdot \;\middle|\; \mathbf{X},\, \mathbf{\tilde{q}}\right), \quad 
\mathbf{\tilde{q}} = [\mathbf{q}, \mathbf{v}_{\mathrm{help}}],
\]  
where recall \(\mathbf{v}_{\mathrm{help}}\) is an instruction for producing helping knowledge (cf. Section \ref{method:early}). This mismatch induces a \emph{task shift}: the TSLM $\pi^{T}$, trained to produce direct answers, tends to generate hallucinated content rather than faithful, unbiased analysis. As a result, \(\mathbf{K}^{T}\) is systematically misaligned with the desired ground-truth knowledge for injection. Constructing large expert-annotated corpora specifically for this injection setting could mitigate the issue but is prohibitively costly.

\paragraph{Thinking Transfer.}
To resolve the task shift between answering and supplying knowledge, we propose to align the specialist with its injection role by training it to produce a thinking trace before any answer. Then, such a specialist thinking trace is served directly as the knowledge source, 
\begin{align}
\label{eq:TSLM_think}
\mathbf{K}^{T}_{think} \;:=\; \mathbf{r}^{T}  \;\sim\; \pi^{T}\!\left(\,\cdot \;\middle|\; \mathbf{X},\, \mathbf{q}\right).
\end{align}
At inference, we perform injection by starting GRLM reasoning with this analysis and a brief reflection cue,
\begin{align}
\label{eq:early_injection}
\mathbf{r}^{\mathrm{Inj}}_{\le 1}
=
\mathsf{Inject}_{reflect}\!\big(\,\emptyset,\, \mathbf{r}^{T}\big)
=
\big[\, \mathbf{r}^{T},\, \boldsymbol{v}_{\mathrm{reflect}}\,\big],
\end{align}
and then continue the general reasoning process. This design naturally aligns training and deployment: the TSLM learns to produce analysis first, and the injected analysis serves as a grounded knowledge source for steering the reasoner.

\paragraph{RL Training without Thinking Supervision.}
Directly training a TSLM to produce analysis-first thinking traces, as in Eq. (\ref{eq:TSLM_think}), is challenging as most time-series diagnostic datasets contain only ground-truth answers \(\mathbf{y}^*\) but not the intermediate reasoning traces \(\mathbf{r}^*\). To overcome this, we employ reinforcement learning with verifiable rewards (RLVR) \citep{guo2025deepseek}. Let $\mathbf{z}=[\mathbf{r},\mathbf{y}]$ denote a sampled completion containing both a trace and an answer. 
For each context $(\mathbf{X},\mathbf{q})$, we draw a group of $G$ completions $\{\mathbf{z}_i\}_{i=1}^G$ and optimize with the group-relative objective: 
\begin{align}
\label{eq:grpo_main}
\max_{\theta}\; \mathbb{E}_{\{\mathbf{z}_i\}_{i=1}^G \sim \pi_{\theta}(\cdot\mid \mathbf{X},\mathbf{q})}
\Bigl[\mathcal{L}_{GRPO}\bigl(\theta, \{R(\mathbf{z}_i)\}_{i=1}^G \bigr)\Bigr],
\end{align}
with $R(\mathbf{z}) \;=\; r_{\mathrm{fmt}}(\mathbf{z}) + r_{\mathrm{hard}}(\mathbf{z})$, where \(r_{\mathrm{fmt}}\in\{0,1\}\) is a format reward that equals 1 if the output follows the target structure  
\[
\langle\mathrm{think}\rangle\, \mathbf{r}\, \langle/\mathrm{think}\rangle \;
\langle\mathrm{answer}\rangle\, \mathbf{y}\, \langle/\mathrm{answer}\rangle,
\]  
and 0 otherwise. The hard reward \(r_{\mathrm{hard}}\in\{0,1\}\) equals 1 if the predicted answer \(\mathbf{y}\) matches the ground-truth \(\mathbf{y}^*\), and 0 otherwise. Here the objective is computed over groups of G sampled completions, with rewards normalized within the group; the detailed form of $\mathcal{L}_{GRPO}$ is provided in Appendix~\ref{app:grpo}. Importantly, \emph{no labeled traces are required}: the policy is driven to \emph{elicit} analysis-first reasoning purely through structural and correctness feedback. This is particularly valuable for time-series diagnostics, where ground-truth outcomes are available but the intermediate causal links between the time seires $\mathbf{X}$ and the underlying root cause $\mathbf{y}^*$ is unobserved and must be discovered through learning. The full algorithm is summarized in Algorithm \ref{alg:ki-rl}.

\section{Benchmark: SenTSR-Bench}

\begin{table}[htbp]
\centering
\caption{Comparison of time-series diagnostic reasoning benchmarks. }
\label{tab:benchmark-comparison}
\resizebox{\linewidth}{!}{
\begin{tabular}{lcccc}
\toprule
Benchmark & \makecell{New \\ Time-Series?} & \makecell{Real-\\World?} & \makecell{Multi-stage Advancing\\Questions?} & \makecell{Anno-\\tation?} \\
\midrule
TSEvol \citep{xie2024chatts}    & \loss{\xmark} & \gain{\cmark}& \loss{\xmark} & LLM \\
TS\&Language \citep{merrill2024language}       & \gain{\cmark} & \loss{\xmark} & \loss{\xmark} & LLM \\
MTBench \citep{chen2025mtbench}  & \gain{\cmark} & \gain{\cmark} & \loss{\xmark} & LLM \\
Sensor-TSR (Ours)  & \gain{\cmark}& \gain{\cmark}& \gain{\cmark}& Human \\
\bottomrule
\end{tabular}}
\end{table}

\begin{wrapfigure}{r}{0.5\textwidth}
    \centering
    \includegraphics[width=0.48\textwidth]{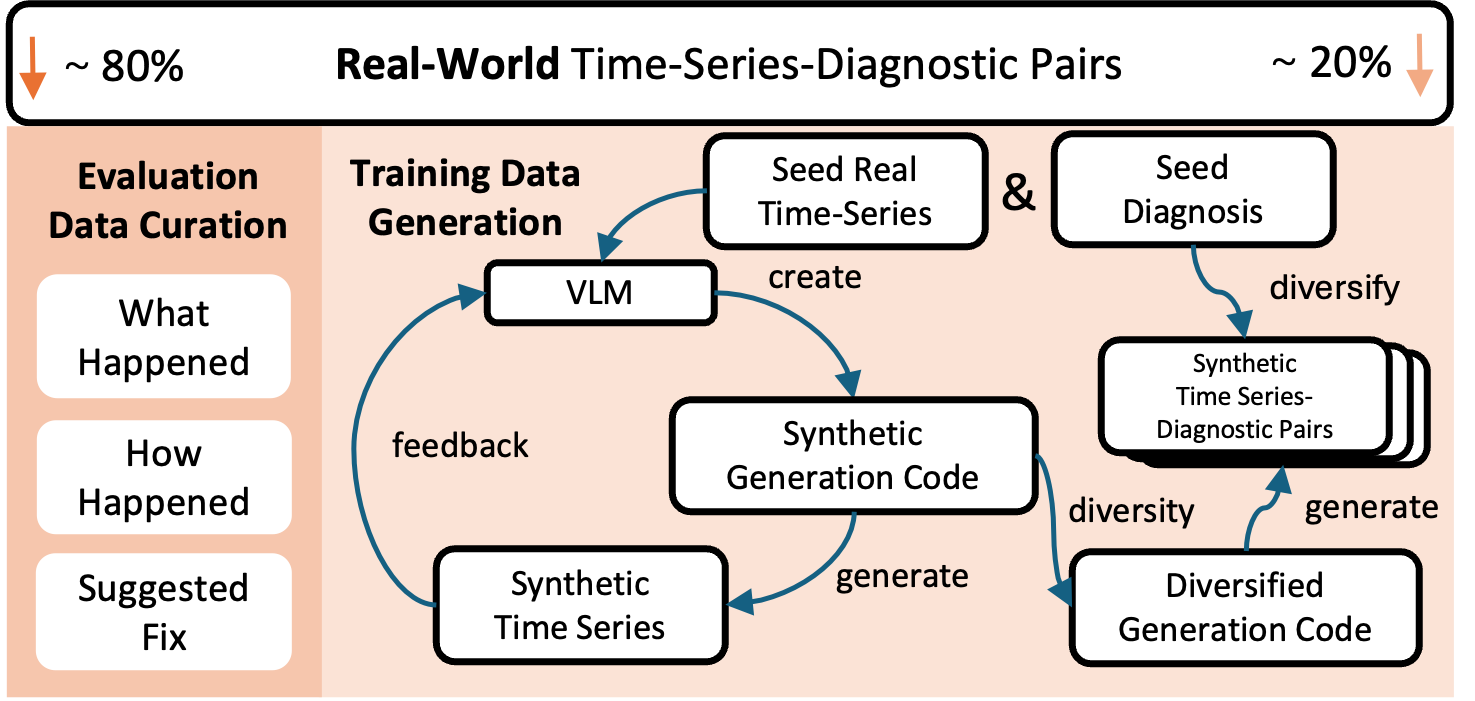}
    \caption{SenTSR-Bench Construction pipeline.}
    \label{fig:benchmark_workflow}
\end{wrapfigure}

Despite the growing interest in time-series diagnostic reasoning, there are still very limited high-quality datasets that couple real-world time-series with textual diagnostic annotations. As summarized in Table~\ref{tab:benchmark-comparison}, existing work primarily rely on LLM-annotated versions of public time-series datasets or fully synthetic time series–text pairs, and typically provide only a single question per series, falling short of capturing real-world diagnostic complexity. In this work, we introduce \emph{SenTSR-Bench}, a new benchmark directly motivated by real-world sensor monitoring for machine breakdown diagnosis and troubleshooting. The benchmark consists of de-identified, multivariate time-series signals collected from vibration (acceleration, velocity) and temperature sensors, paired with human-curated diagnostic annotations.

\textit{SenTSR-Bench} moves beyond anomaly flagging and evaluate the full procedure of diagnostic reasoning. The benchmark contains different levels of questions: (i) what happened (recognizing anomalous segments in multivariate time-series), (ii) how happened (inferring plausible root causes behind the observed signals), and (iii) suggested fix (proposing potential corrective actions). This benchmark provides a realistic and challenging testbed for developing models capable of robust, context-aware diagnostic reasoning. Figure \ref{fig:benchmark_workflow} shows a simplified version of the data construction pipeline. Additional details on the benchmark construction pipeline is provided in Appendix \ref{app:dataset}.

\paragraph{Evaluation Dataset Curation}
To build the evaluation dataset, we follow a three-stage curation pipeline. 
First, we filter 110 multivariate sensor streams out of an initial pool of over 2,000 candidate samples, selecting those that exhibit clear anomalous patterns tied to potential troubleshooting actions. All signals are then standardized to remove sensitive information. 
Second, we design an annotation pipeline that generates multi-stage diagnostic text while preserving privacy, producing faithful but de-identified annotations. 
Third, we construct 330 multiple-choice questions (MCQ) by pairing ground-truth answers with distractors. 
This process yields a benchmark that is both realistic and privacy-preserving, while supporting rigorous evaluation of anomaly recognition, root cause reasoning, and fix proposal tasks.

\paragraph{Training Dataset Generation.}
A key challenge is generating diverse \textit{multivariate} sensor streams with a small number of real-world seeds are available. To address this, we design a two-stage synthetic generation pipeline powered by vision–language models (VLMs). \emph{Stage 1: Iterative code synthesis} prompts a VLM with plots and context from 23 de-identified seeds to produce Python codes that mimic the original behaviors. \emph{Stage 2: Diversification and simplification} transforms these simulators into compact stochastic generators that introduce randomized dynamics and parameter variation, yielding broad families of realistic synthetic series. The resulting synthetic data are then used to construct 6,000 MCQ training entries consistent with the evaluation design. 

\section{Experiments}

\subsection{Experiment Setup}
\paragraph{Datasets.}
For evaluation, we use \textit{SenTSR-Bench}, our de-identified, real-world benchmark of multivariate time-series with three progressively harder tasks: \emph{What happened} (key time-series anomaly characterization), \emph{How it happened} (root-cause diagnosis), and \emph{Suggested fix} (action recommendation). We additionally assess the performance on two public benchmarks: \textit{TSEvol} (Dataset A) from \cite{xie2024chatts}, which covers \emph{inductive}, \emph{deductive}, and \emph{causal} reasoning, and \emph{MCQ2} dataset from \textit{TS\&Language} Benchmark \citep{merrill2024language}, which poses relational queries over paired time-series under textual context. Additional details on the datasets are provided in Appendix \ref{app:dataset}.

\paragraph{Implementation and Evaluation} For the general reasoning model, we test the open-source models \texttt{DeepSeekR1-}\texttt{Distilled-}\texttt{Qwen‐32B} \citep{guo2025deepseek} and \texttt{Qwen3-32B} \citep{yang2025qwen3} as well as closed-source models \texttt{Claude3.7} \citep{anthropic2025claude37sonnet} with time-series encoded as either the vision form (\texttt{-vision}) or the textual form (\texttt{-text}). All models are set up with standard config. For fine-tuned TSLM, we primarily use \texttt{Qwen2.5-VL-3B} \citep{bai2025qwen25vl} for SFT and RL training. We also use \texttt{ChatTS-14B} \citep{xie2024chatts} for injection design exploration. For evaluation, generative QA tasks (inductive reasoning in \textit{SenTSR-Bench}) are evaluated using RAGAS. Verifiable tasks report \emph{accuracy}.  All results are averaged over three independent runs. Further details on implementation and evaluation are provided in Appendix \ref{app:implementation}. 

\begin{table*}[t]
\centering
\caption{Reasoning performance on \emph{SenTSR-Bench} Benchmark (mean$\pm$std). Best per block are \textbf{bolded}. The last two columns report relative gains (in \%) for Injection rows vs.\ the corresponding specialized TSLM and the zero-shot general reasoner (GRLM), respectively.}
\label{tab:mcq-sft-rl}
\resizebox{0.85\linewidth}{!}{
\begin{tabular}{llccc|crr}
\toprule
Model & Paradigm & What Happened & How Happened & Suggested Fix & \textbf{Overall} & \multicolumn{2}{c}{\textbf{Improvement vs.}} \\
\cmidrule(lr){7-8}
&&&&& & \textbf{TSLM} & \textbf{GRLM} \\
\midrule
TSLM (Qwen-VL-3B) & SFT & 0.530 $\pm$ 0.037 & 0.567 $\pm$ 0.029 & {0.548 $\pm$ 0.011} & 0.549 $\pm$ 0.019 & — & — \\
& RL  & 0.512 $\pm$ 0.038 & 0.594 $\pm$ 0.019 & 0.546 $\pm$ 0.009 & 0.551 $\pm$ 0.014 & — & — \\
GRLM (Claude3.7-Text) & Zero-shot          & 0.712 $\pm$ 0.019 & 0.409 $\pm$ 0.033 & 0.473 $\pm$ 0.024 & 0.531 $\pm$ 0.011 & — & — \\
& Few-shot           & 0.691 $\pm$ 0.009 & 0.561 $\pm$ 0.011 & 0.509 $\pm$ 0.009 & 0.587 $\pm$ 0.006 & — & \gain{+10.5\%} \\
\rowcolor{baselinegray}
TSLM + GRLM & SFT-Injection & {0.742 $\pm$ 0.023} & {0.603 $\pm$ 0.021} & \textbf{0.558 $\pm$ 0.019} & {0.634 $\pm$ 0.006} & \gain{+15.5\%} & \gain{+19.4\%} \\
\rowcolor{baselinegray}
& RL-Injection  & \textbf{0.779 $\pm$ 0.014} & \textbf{0.627 $\pm$ 0.018} & 0.542 $\pm$ 0.028 & \textbf{0.650 $\pm$ 0.010} & \gain{+18.0\%} & \gain{+22.4\%} \\
\addlinespace[1em]
TSLM (Qwen-VL-3B) & SFT & 0.530 $\pm$ 0.037 & 0.567 $\pm$ 0.029 & 0.548 $\pm$ 0.011 & 0.549 $\pm$ 0.019 & — & — \\
& RL  & 0.512 $\pm$ 0.038 & {0.594 $\pm$ 0.019} & 0.546 $\pm$ 0.009 & 0.551 $\pm$ 0.014 & — & — \\
GRLM (Claude3.7-Vision) & Zero-shot          & 0.764 $\pm$ 0.016 & 0.542 $\pm$ 0.019 & 0.555 $\pm$ 0.018 & 0.620 $\pm$ 0.006 & — & — \\
& Few-shot           & {0.824 $\pm$ 0.014} & 0.552 $\pm$ 0.014 & 0.555 $\pm$ 0.018 & 0.643 $\pm$ 0.005 & — & \gain{+3.7\%} \\
\rowcolor{baselinegray}
TSLM + GRLM & SFT-Injection & 0.756 $\pm$ 0.031 & 0.588 $\pm$ 0.013 & \textbf{0.649 $\pm$ 0.029} & {0.665 $\pm$ 0.020} & \gain{+21.1\%} & \gain{+7.3\%} \\
\rowcolor{baselinegray}
& RL-Injection  & \textbf{0.827 $\pm$ 0.009} & \textbf{0.661 $\pm$ 0.014} & {0.597 $\pm$ 0.032} & \textbf{0.695 $\pm$ 0.012} & \gain{+26.1\%} & \gain{+12.1\%} \\
\bottomrule
\end{tabular}}
\end{table*}

\begin{table*}[t]
\centering
\caption{Reasoning performance on \textit{TSEvol} and \textit{TS\&Language} Benchmark (mean$\pm$std). Best per block are \textbf{bolded}. The last two columns report relative gains (in \%) for Injection rows vs.\ the corresponding specialized TSLM and the zero-shot general reasoner (GRLM), respectively.}
\label{tab:rl-thinking}
\resizebox{\linewidth}{!}{
\begin{tabular}{llcccc|crr}
\toprule
Model & Paradigm & Causal & Deductive & Inductive & MCQ2 & \textbf{Overall} & \multicolumn{2}{c}{\textbf{Improvement vs.}} \\
\cmidrule(lr){8-9}
&&&&&& & \textbf{TSLM} & \textbf{GRLM} \\
\midrule
TSLM (Qwen-VL-3B) & SFT & {0.623 $\pm$ 0.006} & {0.520 $\pm$ 0.013} & 0.357 $\pm$ 0.010 & {0.507 $\pm$ 0.032} & 0.502 $\pm$ 0.005 & — & — \\
& RL  & \textbf{0.627 $\pm$ 0.016} & 0.496 $\pm$ 0.014 & 0.313 $\pm$ 0.023 & \textbf{0.597 $\pm$ 0.031} & 0.508 $\pm$ 0.006 & — & — \\
GRLM (Qwen3-32B) & Zero-shot & 0.507 $\pm$ 0.041 & 0.473 $\pm$ 0.035 & \textbf{0.623 $\pm$ 0.036} & 0.407 $\pm$ 0.015 & 0.502 $\pm$ 0.023 & — & — \\
& Few-shot  & 0.622 $\pm$ 0.028 & 0.473 $\pm$ 0.035 & 0.460 $\pm$ 0.033 & 0.427 $\pm$ 0.015 & 0.495 $\pm$ 0.010 & — & \loss{-1.4\%} \\
\rowcolor{baselinegray}
TSLM+GRLM & SFT-Injection & 0.569 $\pm$ 0.035 & \textbf{0.543 $\pm$ 0.013} & {0.592 $\pm$ 0.031} & 0.410 $\pm$ 0.036 & {0.528 $\pm$ 0.008} & \gain{+5.2\%} & \gain{+5.2\%} \\
\rowcolor{baselinegray}
& RL-Injection  & \textbf{0.627 $\pm$ 0.025} & 0.512 $\pm$ 0.047 & 0.588 $\pm$ 0.035 & 0.490 $\pm$ 0.046 & \textbf{0.554 $\pm$ 0.021} & \gain{+9.1\%} & \gain{+10.4\%} \\
\addlinespace[1em]

TSLM (Qwen-VL-3B) & SFT & 0.623 $\pm$ 0.006 & {0.520 $\pm$ 0.013} & 0.357 $\pm$ 0.010 & {0.507 $\pm$ 0.032} & 0.502 $\pm$ 0.005 & — & — \\
& RL  & {0.627 $\pm$ 0.016} & 0.496 $\pm$ 0.014 & 0.313 $\pm$ 0.023 & \textbf{0.597 $\pm$ 0.031} & 0.508 $\pm$ 0.006 & — & — \\
GRLM (R1-Distilled-Qwen-32B) & Zero-shot  & 0.522 $\pm$ 0.022 & {0.550 $\pm$ 0.054} & {0.525 $\pm$ 0.015} & 0.483 $\pm$ 0.015 & 0.520 $\pm$ 0.010 & — & — \\
& Few-shot   & 0.542 $\pm$ 0.017 & \textbf{0.558 $\pm$ 0.040} & 0.478 $\pm$ 0.022 & 0.513 $\pm$ 0.021 & 0.523 $\pm$ 0.007 & — & \gain{+0.6\%} \\
\rowcolor{baselinegray}
TSLM+GRLM & SFT-Injection & 0.594 $\pm$ 0.023 & 0.535 $\pm$ 0.023 & 0.519 $\pm$ 0.004 & 0.490 $\pm$ 0.020 & {0.534 $\pm$ 0.007} & \gain{+6.4\%} & \gain{+2.7\%} \\
\rowcolor{baselinegray}
& RL-Injection  & \textbf{0.634 $\pm$ 0.013} & 0.543 $\pm$ 0.013 & \textbf{0.532 $\pm$ 0.010} & {0.537 $\pm$ 0.032} & \textbf{0.561 $\pm$ 0.011} & \gain{+10.4\%} & \gain{+7.9\%} \\
\bottomrule
\end{tabular}}
\end{table*}

\begin{table*}[ht]
\centering
\caption{{Performance with different injection strategy on \textit{TSEvol} and \textit{TS\&Language} Benchmark (mean$\pm$std).}  Best results are \textbf{bolded}.}
\label{TAB_TSEvol_MCQ}
\resizebox{\linewidth}{!}{
\begin{tabular}{llcccc|c}
\toprule
Model & Injection Strategy & Inductive & Deductive & Causal & MCQ2 & \textbf{Overall} \\
\midrule
TSLM (ChatTS-14B) & — & {0.812 $\pm$ 0.007} & 0.597 $\pm$ 0.013 & \textbf{0.732 $\pm$ 0.006} & 0.590 $\pm$ 0.026 & 0.683 $\pm$ 0.010 \\
GRLM (Claude3.7-Text) & — &
0.763 $\pm$ 0.021 & 0.612 $\pm$ 0.029 & 0.645 $\pm$ 0.021 & 0.640 $\pm$ 0.014 & 0.665 $\pm$ 0.010 \\
\rowcolor{baselinegray}
TSLM + GRLM& Intermediate &
0.805 $\pm$ 0.026 & {0.659 $\pm$ 0.022} & 0.645 $\pm$ 0.010 & \textbf{0.703 $\pm$ 0.037} & 0.703 $\pm$ 0.006 \\
\rowcolor{baselinegray}
& Late &
0.791 $\pm$ 0.014 & \textbf{0.667 $\pm$ 0.011} & {0.703 $\pm$ 0.019} & 0.680 $\pm$ 0.022 & {0.710 $\pm$ 0.003} \\
\rowcolor{baselinegray}
& Early &
\textbf{0.824 $\pm$ 0.019} & 0.643 $\pm$ 0.011 & {0.703 $\pm$ 0.019} & {0.690 $\pm$ 0.016} & \textbf{0.715 $\pm$ 0.003} \\
\addlinespace[1em]

TSLM (ChatTS-14B) & — & {0.812 $\pm$ 0.007} & 0.597 $\pm$ 0.013 & {0.732 $\pm$ 0.006} & 0.590 $\pm$ 0.026 & 0.683 $\pm$ 0.010 \\
GRLM (Claude3.7-Vision) & — &
0.792 $\pm$ 0.016 & 0.643 $\pm$ 0.011 & 0.630 $\pm$ 0.009 & 0.690 $\pm$ 0.008 & 0.689 $\pm$ 0.005 \\
\rowcolor{baselinegray}
TSLM + GRLM& Intermediate &
0.809 $\pm$ 0.011 & {0.674 $\pm$ 0.000} & 0.663 $\pm$ 0.009 & {0.713 $\pm$ 0.017} & 0.715 $\pm$ 0.004 \\
\rowcolor{baselinegray}
& Late &
0.800 $\pm$ 0.019 & \textbf{0.682 $\pm$ 0.011} & 0.707 $\pm$ 0.009 & 0.697 $\pm$ 0.005 & {0.721 $\pm$ 0.005} \\
\rowcolor{baselinegray}
& Early &
\textbf{0.825 $\pm$ 0.011} & 0.643 $\pm$ 0.029 & \textbf{0.746 $\pm$ 0.005} & \textbf{0.730 $\pm$ 0.014} & \textbf{0.736 $\pm$ 0.002} \\
\bottomrule
\end{tabular}}
\end{table*}

\subsection{Performance Analysis}

For performance analysis, we evaluate the proposed knowledge injection framework on both our newly released \textit{SenTSR-Bench} benchmark and  public benchmarks. We test injection across different TSLM training and injection paradigms (SFT/RL-based), and multiple general reasoning models. Results are presented in Table~1. Here are the observations:

\paragraph{Injection Lifts Both Baselines.}
Across all benchmark datasets, injecting TSLM knowledge, whether from SFT or RL-tuned TSLM, consistently boosts accuracy over both stand-alone specialists and stand-alone reasoners. On \textit{SenTSR-Bench}, gains range from {+15.5\% to +26.1\%} over the specialized TSLM and {+7.3\% to +22.4\%} over the general GRLM; improvements span all three tasks and are most pronounced on \emph{How happened}, which involves both in-domain anomaly detection knowledge and strong causal reasoning capacity. On the public benchmarks, we observe similar trends: {+5.2\% to +10.4\%} over the specialist and {+2.7\% to +10.4\%} over the reasoner. The injected variant shows robustness: even when the TSLM performs poorly (e.g., the \emph{Inductive} task), the injected model leverages the reasoner’s critical thinking capacity to maintain competitive performance. Taken together, injection delivers the best \textit{overall} performance across settings.

\paragraph{RL-based  Injection Consistently Yields Larger Gains.}
Compared with SFT-based injection, RL-based \emph{thinking transfer} delivers consistently larger improvements over zero-shot GRLMs: when measuring gains, RL-based injection provides {1.66}\(\times\) the improvement on \texttt{Claude3.7-Vision}, {2.00}\(\times\) on \texttt{Qwen3-32B}, and {2.92}\(\times\) on \texttt{DeepSeekR1-Distilled-Qwen-32B}. Both SFT and RL injection outperform few-shot prompting, but RL provides the biggest lifts (e.g., {3.27}\(\times\) than few-shot  on \texttt{Claude3.7-Vision}). Moreover, injection is more \emph{token-efficient}: while tokenized multivariate time-series in \textit{TSevol} can exceed \(\sim\!50\mathrm{k}\) tokens, making few-shot prompts infeasible, injection instead provides a compact analysis snippet through thinking prefill, offering a more scalable  mechanism for time-series diagnostic reasoning.

\begin{wrapfigure}{r}{0.5\textwidth}
    \vspace{-12pt}
    \centering
    \includegraphics[width=0.48\textwidth]{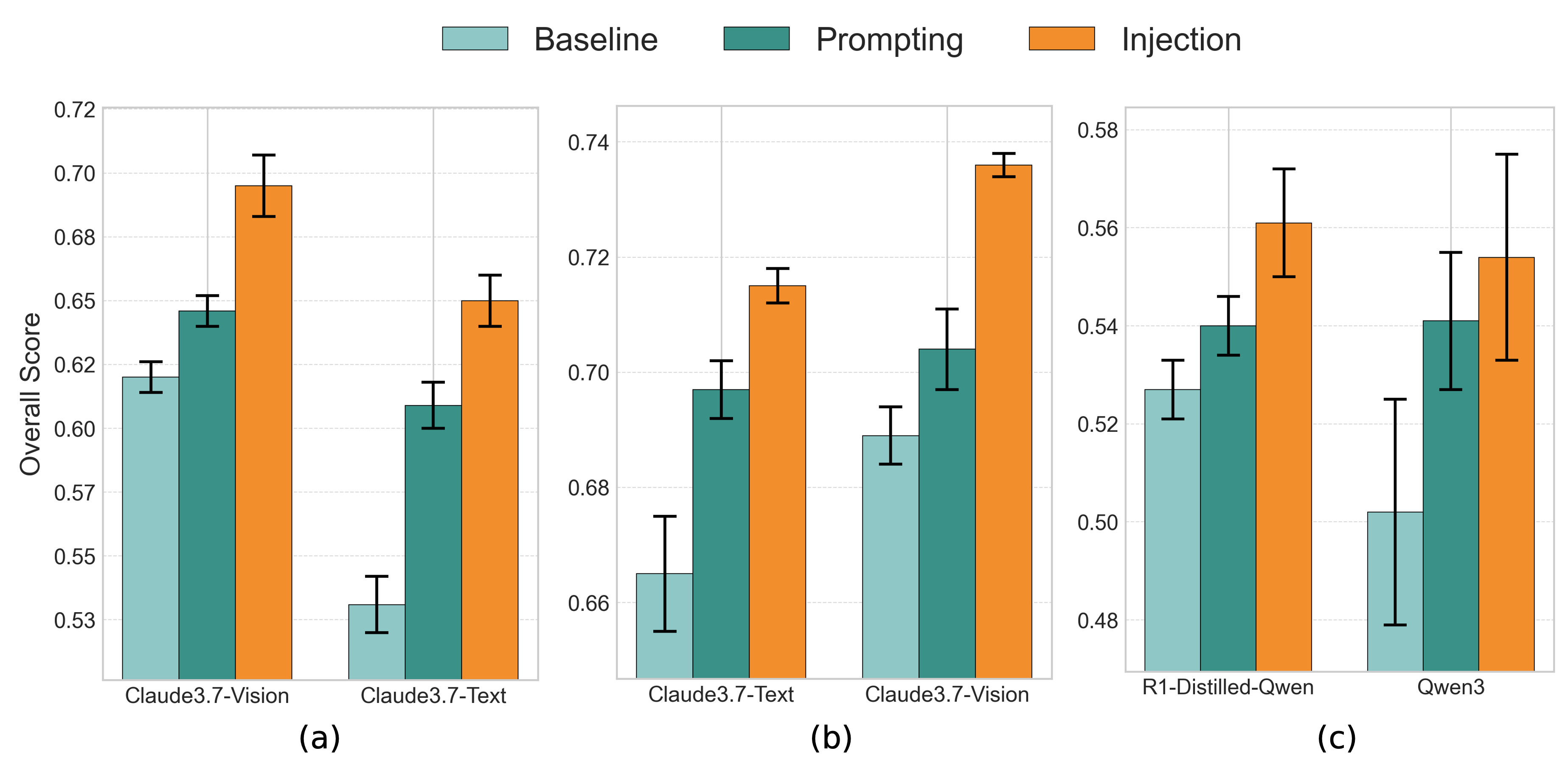}
    \caption{Comparison of baseline (zero-shot) reasoning, knowledge prompting, and knowledge injection. (a) \emph{SenTSR-Bench} Benchmark with Qwen-VL-3B (RL) as the TSLM. (b) \emph{TSEvol} and \emph{TS\&Language} Benchmarks with Qwen-VL-3B (RL) as the TSLM. (c) \emph{TSEvol} and \emph{TS\&Language} Benchmarks with ChatTS-14B as the TSLM. Across all settings, the injection-based method consistently outperforms others.}
    \label{fig:prompt-vs-inject}
    \vspace{-10pt}
\end{wrapfigure}

\subsection{Framework Analysis}
\label{sec:framework_analysis}

\paragraph{Comparison across Different Injection Strategies.} We evaluate three \emph{injection strategies}—\emph{early}, \emph{intermediate}, and \emph{late}.  \emph{Early injection} inserts the specialist’s analysis immediately after the opening token, \emph{Intermediate injection} with correcting the lowest-confidence token position and \emph{Late injection} appends a specialist-generated critique to the full reasoning trace at the end. Further implementation details are provided in Appendix \ref{app:implementation}. We use \texttt{ChatTS-14B} \citep{xie2024chatts} here (rather than smaller specialists) as it is tuned with a broad range of time-series QA tasks and thus better trained to investigate the problem. Table~3 reports results across \texttt{Claude3.7-Text} and \texttt{Claude3.7-Vision}. Results show that all three strategies consistently outperform both baselines, aligned with our previous findings. Among them, \emph{early injection} yields the strongest gains across both text and vision reasoners. One key reason is that mid/late injection requires the specialist to read and revise long reasoning traces, which lie outside the distribution of QA-style SFT and lead to drift or hallucination. Early injection aligns naturally with the specialist’s strengths of producing short, focused analyses that can be directly prefixed into the reasoning trajectory. 
\paragraph{Comparison between Prompting and Knowledge Injection.} We next compare our knowledge injection approach with a prompting-based alternative. In the prompting setup, the same TSLM outputs are provided to the reasoning model as additional prompt instructions, rather than being integrated into its internal reasoning trace. Figure~\ref{fig:prompt-vs-inject} contrasts the three strategies: baseline (zero-shot) reasoning, prompting, and injection. Across all model families, from open-source to closed-source, and across all three benchmark datasets, we observe that injection consistently outperforms prompting. This advantage arises because injection places domain knowledge directly inside the reasoning process, which encourages the model to interact with and reflect upon the knowledge more effectively. In contrast, when knowledge is only presented as external prompt instructions, the reasoning model often fails to fully incorporate it. See Appendix~\ref{app:inject_over_prompt} for illustrative case studies.

\subsection{Additional Analysis}
We ablate the role of direct time-series access by removing the raw series from the GRLM input, showing that relying solely on the TSLM’s textual summary creates an information bottleneck that limits downstream reasoning (Appendix~\ref{app:ablation_tslm_summary}). We also compare injection against prompting-based alternatives such as few-shot, self-consistency, and tree-of-thought in terms of accuracy and inference latency (Appendix~\ref{app:accuracy_latency}). To assess training data requirements, we measure the sensitivity of TSLM performance to synthetic data diversity (Appendix~\ref{app:sensitivity_diversity}), and to examine whether the gains from injection can be replicated by stronger RL objectives alone, we evaluate DAPO, GSPO, and CISPO alongside GRPO (Appendix~\ref{app:rl_ablation}). Qualitative case studies comparing standalone baselines with injection and contrasting knowledge prompting with knowledge injection are presented in Appendix~\ref{app:case_study}.

\section{Related Work}
Time-series reasoning has recently attracted growing interest. One line of work studies prompting-based structured reasoning over temporal data \citep{jiang2025explainable,liu2025evaluating,merrill2024language,liu2025picture}. Another line develops specialist models post-trained on time-series–text pairs \citep{kong2025time,xie2024chatts}. While these approaches show promise, the former lacks domain-specific priors for capturing key diagnostic patterns, and the latter often overfits to in-domain data and struggles with generalization. Our knowledge-injection framework aims to bridge these gaps by combining the reasoning capacity of general LLMs with domain-aligned insights from time-series specialists. Another related direction investigates interventions on the reasoning process. Prior work has explored modifying reasoning traces or internal reasoning for improved faithfulness, safety, and instruction following \citep{wu2025effectively,arcuschin2025chain,baker2025monitoring} as well as methods for controlling the length of reasoning traces to balance accuracy and efficiency \citep{han2024token,aggarwal2025l1,lee2025well}. Our work differs in that we explicitly inject domain knowledge from a specialized model into a general reasoning model, with a specific focus on diagnostic reasoning over time-series data. See Appendix \ref{app:related_work} for additional related works on time-series reasoning in forecasting, time-series reasoning benchmarks.
\section{Conclusion}  
In this paper, we introduced a \emph{knowledge injection} framework that combines domain knowledge from time-series specialists with the strong reasoning ability of large general LLMs. We further proposed RL-based \emph{thinking transfer} for knowledge injection, which naturally elicits analysis-first traces without supervision, enabling effective and task-aligned injection. In addition, we released \emph{SenTSR-Bench}, a real-world benchmark for time-series diagnostic reasoning with multi-stage questions covering anomaly recognition, root-cause diagnosis, and corrective suggestions. Across \textit{SenTSR-Bench} and public datasets, our injeciton framework achieves 7.9\%–26.1\% improvements over standalone baselines. We encourage exploration on \textit{SenTSR-Bench} and further investigation of knowledge injection approaches for broader time-series diagnostic reasoning tasks.

\bibliographystyle{apalike}
\bibliography{ref}

\clearpage
\appendix
\thispagestyle{empty}

\onecolumn

\section{Additional Technical Details}

\subsection{Details of GRPO Training Objective}
\label{app:grpo}
For completeness, we provide the explicit form of the Group Relative Policy Optimization (GRPO) objective $\mathcal{L}_{GRPO}(\theta, R(\mathbf{z}))$ \citep{guo2025deepseek} used in Eq.~(\ref{eq:grpo_main}).

Given a training context $(\mathbf{X}, \mathbf{q})$, we first sample a group of $G$ complete sequences
\[
\bigl\{ \mathbf{z}_i \bigr\}_{i=1}^G \sim \pi_{\theta_{\mathrm{old}}}(\cdot \mid \mathbf{X}, \mathbf{q}),
\]
where each $\mathbf{z}_i$ contains both a reasoning trace and a final answer. We then compute scalar rewards $\{r_i\}_{i=1}^G$ for each sampled sequence using the composite reward function $R(\mathbf{z})$.  

We normalize the rewards into advantages by subtracting the group mean and dividing by the standard deviation:
\[
\hat{A}_i = \frac{r_i - \mu_r}{\sigma_r}, 
\quad 
\mu_r = \frac{1}{G} \sum_{j=1}^G r_j,
\quad 
\sigma_r = \sqrt{\tfrac{1}{G} \sum_{j=1}^G (r_j - \mu_r)^2 + \gamma},
\]
where $\gamma$ is a small constant to ensure numerical stability.  
Each token $z_{i,k}$ in sequence $\mathbf{z}_i$ shares the same normalized advantage $\hat{A}_i$, ensuring stable gradient updates across contexts.

We then optimize the clipped surrogate objective with KL regularization against a frozen reference model $\pi_{\mathrm{ref}}$:
\begin{align*}
\mathcal{L}_{GRPO}(\theta) 
&= \frac{1}{G} \sum_{i=1}^G \frac{1}{|\mathbf{z}_i|} \sum_{k=1}^{|\mathbf{z}_i|} 
\min\!\bigl(\rho_{i,k} \hat{A}_i,\, 
\mathrm{clip}(\rho_{i,k}, 1-\epsilon, 1+\epsilon) \hat{A}_i \bigr) - \beta \, \mathrm{KL}\!\left[\pi_\theta(\cdot \mid \mathbf{X},\mathbf{q}) \;\|\; \pi_{\mathrm{ref}}(\cdot \mid \mathbf{X},\mathbf{q})\right],
\end{align*}
where
\[
\rho_{i,k} = \frac{\pi_\theta(z_{i,k} \mid z_{i,<k}, \mathbf{X}, \mathbf{q})}{\pi_{\theta_{\mathrm{old}}}(z_{i,k} \mid z_{i,<k}, \mathbf{X}, \mathbf{q})}
\]
is the token-level importance ratio, $\epsilon$ is the PPO clipping threshold, and $\beta$ is the KL regularization coefficient. This objective balances three forces: (i) improving the likelihood of high-reward completions relative to the old policy, (ii) clipping updates to maintain stability, and (iii) penalizing divergence from a reference model to prevent degeneration.

\subsection{Other Injection Paradigms}
Beyond early insertion, the same framework can be applied for other paradigms by changing \emph{where} we place the snippet and \emph{how} we shape the request. For completeness, here we introduce framework of intermediate and late injection.

\paragraph{Intermediate Knowledge Injection}
The general reasoner first drafts a partial or full reasoning trace. Although the trace is token-level in our notation, in experiments we elicit a sentence-level structure by instructing the model to reason step by step, sentence by sentence, starting from observable time-series evidence and then connecting to higher-level conclusions. Formally, we apply a deterministic segmentation operator that groups tokens into sentences,
\[
\hat{\mathbf{r}}^{G} \;=\; [\, \mathbf{s}_1, \mathbf{s}_2, \ldots, \mathbf{s}_L \,], \qquad \mathbf{s}_\ell \in V^{*}.
\]
Along with each sentence \(\mathbf{s}_\ell\), we prompt the model to output a self-reported confidence score \(c_\ell \in [0,1]\). In this paper we allow the reasoner to complete the draft \(\hat{\mathbf{r}}^{G}\) and then select the sentence with the lowest confidence,
\[
\ell^{*} \;=\; \arg\min_{\ell \in \{1,\ldots,L\}} \, \mathsf{Conf}(\mathbf{s}_\ell).
\]
Let \(k^{*}\) be the token index at the start of \(\mathbf{s}_{\ell^{*}}\). We now shape an assistance query that asks the specialist to judge this specific statement against the time series and to provide evidence or a correction,
\[
\tilde{\mathbf{q}} \;=\; \mathsf{Query}_{\mathrm{assist}}\!\big(\mathbf{q},\, \hat{\mathbf{r}}^{G}_{\le k^{*}},\, \mathbf{s}_{\ell^{*}},\, \mathbf{v}_{\mathrm{judge}}\big),
\]
where \(\mathbf{v}_{\mathrm{judge}}\) instructs the specialist to verify whether \(\mathbf{s}_{\ell^{*}}\) is supported by the time-series \(\mathbf{X}\). The specialist returns knowledge
\[
\mathbf{K}^{T} \;\sim\; \pi^{T}\!\left(\,\cdot \,\middle|\, \mathbf{X},\, \tilde{\mathbf{q}} \right).
\]
We then perform injection by rolling back to the insertion point and inserting a brief reflection cue \(\mathbf{v}_{\mathrm{reflect}}\) between the existing trace and the specialist knowledge,
\[
\mathbf{r}^{\mathrm{Inj}}_{\le k^{*}} 
\;=\; \mathsf{Inject}_{\mathrm{assist}}\!\big(\hat{\mathbf{r}}^{G}_{< k^{*}},\, \mathbf{v}_{\mathrm{reflect}},\, \mathbf{K}^{T}\big)
\;=\; \big[\, \hat{\mathbf{r}}^{G}_{< k^{*}},\, \mathbf{v}_{\mathrm{reflect}},\, \mathbf{K}^{T} \,\big].
\]
The reasoner then resumes generation for \(j \ge k^{*}\) conditioned on \(\mathbf{r}^{\mathrm{Inj}}\) and produces the final answer following Eq.~\eqref{eq:answer_generation}. This design targets the least certain statement in the draft and supplies focused, time-series grounded evidence at that point.

\paragraph{Late Knowledge Injection.}
The general reasoner first produces a complete reasoning trace \(\hat{\mathbf{r}}^{G}\). As in the intermediate setting, we elicit a sentence by sentence structure by prompting the model to enumerate observations from the time series before drawing conclusions. Recall that we segment the trace into sentences
\[
\hat{\mathbf{r}}^{G} \;=\; [\, \mathbf{s}_1, \mathbf{s}_2, \ldots, \mathbf{s}_L \,], \qquad \mathbf{s}_\ell \in V^{*},
\]
where each \(\mathbf{s}_\ell\) states an observation or an intermediate claim about \(\mathbf{X}\). In late injection there is no confidence monitoring. Instead, we submit the entire draft to the specialist for a structured critique. We shape a critique query that includes the question, the full draft, and a critique instruction,
\[
\tilde{\mathbf{q}} \;=\; \mathsf{Query}_{\mathrm{critique}}\!\big(\mathbf{q},\, \hat{\mathbf{r}}^{G},\, \mathbf{v}_{\mathrm{critique}}\big),
\]
where \(\mathbf{v}_{\mathrm{critique}}\) asks the specialist to examine each sentence \(\mathbf{s}_\ell\) against \(\mathbf{X}\), indicate whether it is supported or contradicted, explain why, and if incorrect provide a corrected statement with channel and time references. The specialist returns a knowledge sequence
\[
\mathbf{K}^{T} \;\sim\; \pi^{T}\!\left(\,\cdot \;\middle|\; \mathbf{X},\, \tilde{\mathbf{q}} \right),
\]
which we structure as a list of per sentence judgments and corrections. We then inject after the full draft by appending a brief reflection cue followed by the specialist critique,
\[
\mathbf{r}^{\mathrm{Inj}}
\;=\;
\mathsf{Inject}_{\mathrm{critique}}\!\big(\hat{\mathbf{r}}^{G},\, \mathbf{v}_{\mathrm{reflect}},\, \mathbf{K}^{T}\big)
\;=\;
\big[\, \hat{\mathbf{r}}^{G},\, \mathbf{v}_{\mathrm{reflect}},\, \mathbf{K}^{T} \,\big].
\]
The reasoner performs a short refinement pass that summarizes the critique, reconciles disagreements, and updates its conclusion, then generates the final answer following Eq.~\eqref{eq:answer_generation}. This late insertion supplies broad, time series grounded feedback on the entire draft and encourages reflection before finalization.

\section{Additional Experiment Results}
\begin{figure}
    \centering
    \includegraphics[width=\linewidth]{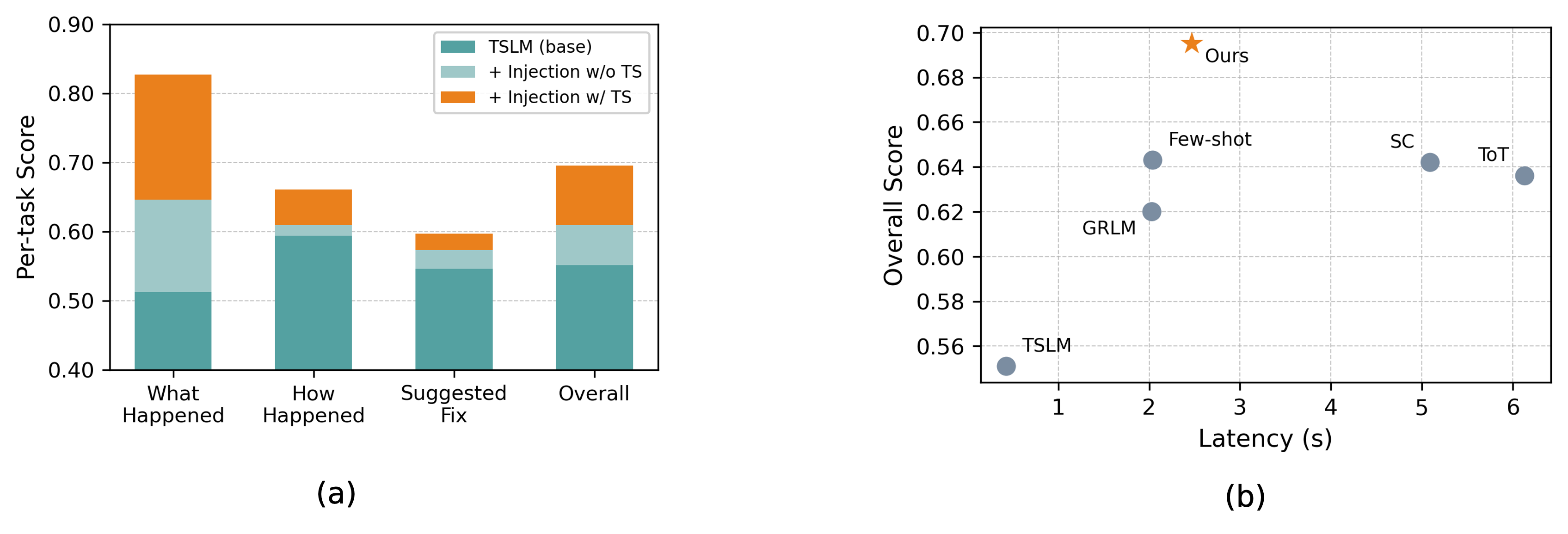}
    \caption{(a) Performance comparison between (i) the standalone TSLM, (ii) knowledge injection where the GRLM receives only the TSLM textual summary (Injection w/o TS), and (iii) full knowledge injection where the GRLM receives both the raw time series and the injected summary (Injection w/ TS). (b) Comparison of overall diagnostic accuracy versus inference latency for different methods.}
    \label{fig:additional_exp}
\end{figure}

\subsection{Ablation on Reliance on TSLM Textual Summaries}
\label{app:ablation_tslm_summary}
We examine whether the GRLM can rely solely on the TSLM’s textual summary, or whether its own direct access to the raw time series $\mathbf{X}$ is necessary for effective reasoning. In our full design, both the TSLM and the GRLM receive $\mathbf{X}$, and the injected summary serves as auxiliary guidance rather than the only information source. This is formalized in Eq. (\ref{eq:GRLM}) and Algorithm \ref{alg:ki-rl}, where the GRLM conditions on $(\mathbf{X}, \mathbf{q}, \mathbf{r})$. The motivation is to avoid a potential failure mode where errors or omissions in the TSLM summary become a single point of failure for downstream reasoning.

To explicitly test this concern, we introduce an ablation in which the GRLM receives only the TSLM-generated textual summary, without access to the raw time series. As shown in Figure \ref{fig:additional_exp} (a), the “Injection w/o TS” variant improves over the standalone TSLM, indicating that transferring learned knowledge from the specialist is beneficial. However, it consistently underperforms the full injection setting. On average, relying only on the textual summary yields approximately a 7\% improvement, whereas full injection with direct time-series access achieves around a 17\% improvement. The gap is most pronounced for the “What Happened” stage, where accurate perception of temporal patterns and anomalies is critical. This ablation demonstrates that the dual-input design is essential: the GRLM does not blindly inherit the TSLM’s errors, but instead combines its own perception of the time series with injected domain knowledge, leading to more robust and accurate diagnostic reasoning.

\subsection{Accuracy and Latency Comparison Across Prompting-Based Alternatives and Injection}
\label{app:accuracy_latency}

We compare our injection-based approach against several commonly used prompting alternatives, including few-shot prompting, self-consistency \citep{wang2022self}, and tree-of-thought \citep{yao2023tree}. Self-consistency is implemented with three independent reasoning runs, and tree-of-thought uses three parallel branches. As shown in Figure \ref{fig:additional_exp} (b), these methods consistently improve over the zero-shot GRLM baseline, confirming that structured prompting and sampling-based reasoning can enhance performance. However, all prompting-based approaches remain noticeably below the injection method in terms of final accuracy, despite incurring substantially higher inference latency. This indicates that the advantage of injection stems from transferring knowledge learned by the TSLM through training, rather than from prompt-level heuristics.

\begin{figure}
    \centering
    \includegraphics[width=\linewidth]{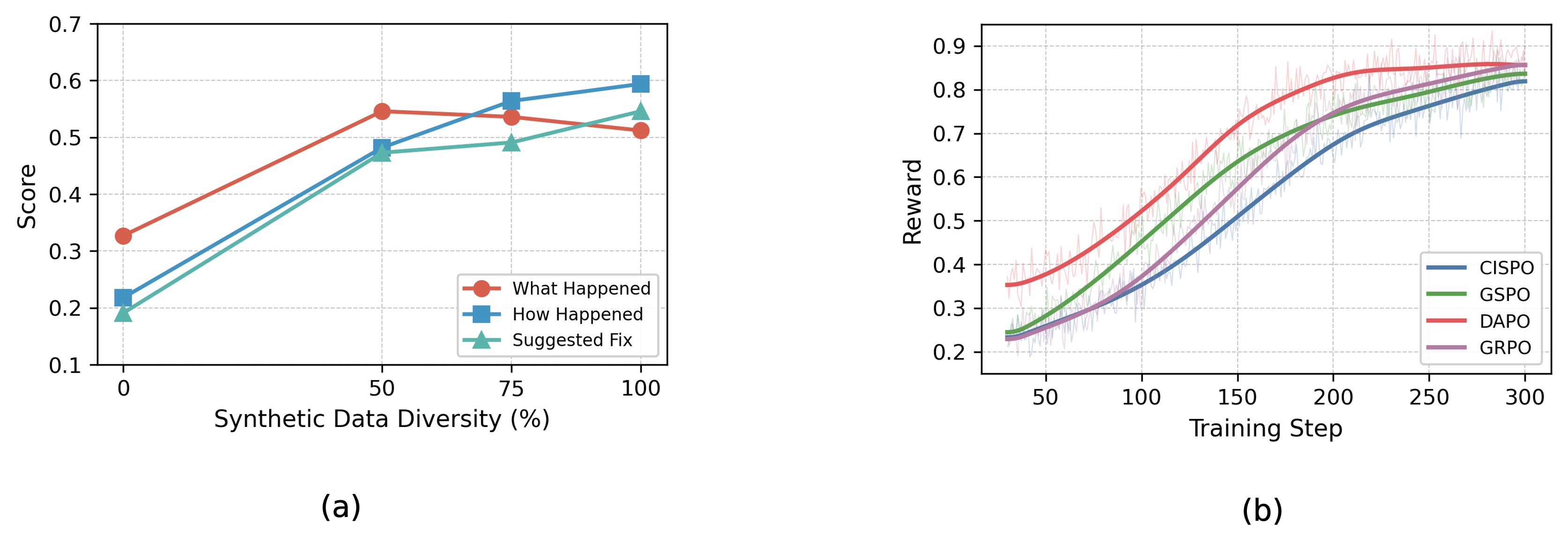}
    \caption{(a) Performance of the TSLM versus synthetic training data diversity, measured by varying the proportion of seed-generated synthetic data used during training. (b) Comparison of reward trajectories for four RL objectives (GRPO, DAPO, GSPO, CISPO) used to train the TSLM.}
    \label{fig:additional_ablation}
\end{figure}

\subsection{Sensitivity of TSLM performance to synthetic data diversity}
\label{app:sensitivity_diversity}

To examine the sensitivity of the TSLM to the quality and diversity of synthetic training data, we conduct an ablation where the model is trained with increasing proportions of seed-generated synthetic data, ranging from no synthetic data to the full dataset. As shown in Figure \ref{fig:additional_ablation} (a), training without synthetic data yields performance close to random guessing across all subtasks, indicating that training is essential for establishing basic time-series diagnostic reasoning ability for small models.

Once training is introduced, performance improves rapidly: using roughly 50\% of the seed-generated data already recovers the majority of the final performance, while increasing diversity beyond 75\% yields only marginal additional gains. This trend is consistent across subtasks, with slightly stronger saturation effects for higher-level reasoning tasks (How Happened and Suggested Fix).

\subsection{Reward convergence under different RL optimization methods}
\label{app:rl_ablation}

Motivated by recent work on efficient R1-style fine-tuning \citep{le2025reasoning,yeo2025demystifying}, we further examine whether more advanced RL objectives can improve TSLM training in our setting. We evaluate three representative RL objectives—DAPO \citep{yu2025dapo}, GSPO \citep{zheng2025group}, and CISPO \citep{chen2025minimax}, alongside our GRPO baseline. As shown in Figure \ref{fig:additional_ablation} (b), methods like DAPO yields faster and smoother reward convergence. At the same time, we observe that the final reward achieved by these methods remains similar across objectives. This suggests that, beyond convergence efficiency, overall performance is primarily constrained by the available supervision and the capacity of the model rather than the specific RL methods.

\section{Additional Related Work}
\label{app:related_work}

\paragraph{Multi-modal Time-Series Forecasting Models} 
Recent there are several lines of research that explores a multimodal solution for time-series analysis to incorporate the information from textual data \citep{liu2025can}. Examples include augmenting series with domain-relevant text \citep{jin2023time,liu2025timecma,liu2024autotimes,liu2025calf}, aligning physiological signals with clinical notes \citep{wan2024meit}, linking stock trends with news \citep{wang2024news,tavakoli2025multi}, and incorporating geographic context \citep{chen2024terra}, traffic data \citep{zhang2024bjtt}, or external events \citep{han2024event} for traffic-flow modeling. However, these work mostly focuses on forecasting tasks rather than multi-modal understanding and diagnostic tasks.

\paragraph{Time-Series Reasoning Models and Benchmarks.}
Time series reasoning has recently drawn growing interest as research moves from prediction toward explanation and diagnosis. Several works explore prompting based reasoning over temporal data \citep{jiang2025explainable,liu2025evaluating,merrill2024language}. \textit{VLTime} \citep{liu2025picture} represents time series as visual plots and queries multimodal models such as GPT4o for zero or few shot interpretation, while \textit{TimeMQA} \citep{kong2025time} formulates question answering tasks using multiple choice reasoning. Both works introduce accompanying benchmarks, \textit{TimerBench} and \textit{TimeMQA}, which are derived from forecasting, classification, or anomaly detection datasets rather than from diagnostic annotations. 
Recent datasets such as \textit{TS\&Language} \citep{merrill2024language} and \textit{TSEvol} \citep{xie2024chatts} extend the setting to textual question–answer tasks, but their explanations are automatically generated by large language models and lack verified diagnostic grounding.
Our \textit{TSRIndustrial} benchmark differs in two key aspects: it provides human-verified diagnostic annotations and introduces a multi-stage problem structure that progresses from identifying anomalies to inferring root causes and suggesting fixes, reflecting the reasoning depth required in real-world maintenance scenarios.
Concurrently, \citet{cao2026more} provide one of the first systematic investigations of LLMs on structured temporal reasoning, focusing on time interval prediction. Their findings reveal that LLMs outperform lightweight statistical baselines yet consistently underperform dedicated machine learning models, and that incorporating additional context does not always improve and can even degrade prediction quality. This formal characterization of LLM temporal reasoning capabilities lays important groundwork for the broader time-series reasoning direction, and extending such analysis beyond interval prediction to diagnostic reasoning remains an interesting future direction.

\section{Dataset Details}
\label{app:dataset}
\subsection{Public Dataset}
We evaluate our framework on two public benchmarks, \textit{TSEvol} and \textit{TSandLanguage}.

\textit{TSEvol} \citep{xie2024chatts} consists of multiple subdatasets, among which we specifically use Dataset A, as it contains real-world time series collected from diverse domains such as AIOps, meteorology, the Numenta Anomaly Benchmark (NAB), and Oracle system metrics. The time series in Dataset A are manually annotated to mark key temporal behaviors, while the contextual prompts and root-cause options are generated automatically by LLM. The dataset includes 525 questions spanning three reasoning categories: (i) \emph{inductive reasoning} — summarizing the physical semantics in univariate or multivariate series, (ii) \emph{deductive reasoning} — verifying temporal conditions, and (iii) \emph{causal reasoning} — selecting the most plausible cause under a given textual context.

\textit{TSandLanguage} (MCQ2) \citep{merrill2024language} is an open-source dataset designed for relational and comparison reasoning between two time series under textual context. The time series, questions, and answers are automatically generated by large language models. Following \cite{xie2024chatts}, we focus on its diagnostic-style multiple-choice subset and exclude etiological reasoning and forecasting components that are not aligned with our evaluation objectives, randomly sampling 100 representative questions.

\subsection{SenTSR-Bench}

\textit{SenTSR-Bench} is a de-identified real-world diagnostic reasoning benchmark derived from industrial sensor systems. It contains 110 multivariate time series paired with 330 human-verified diagnostic questions, spanning three progressive reasoning stages: (i) \emph{what happened} — identifying anomalous signals and temporal patterns, (ii) \emph{how it happened} — inferring plausible root causes behind the observed behavior, and (iii) \emph{suggested fix} — proposing potential corrective actions. The dataset captures realistic multivariate temporal reasoning complexity, from signal interpretation to causal and prescriptive reasoning.

\subsubsection{Evaluation Dataset Curation}
\label{app:curation}
The construction of SenTSR-Bench proceeds in three stages:

\textbf{Stage 1: Signal selection and preprocessing.}  
We start from a large pool of approximately 2,000 multivariate sensor time-series collected from real monitoring systems. From these, we identify 110 streams that display clear anomalous behaviors such as persistent deviations, sharp drops or spikes, or sudden shifts in periodicity. Each selected stream is associated with a downstream troubleshooting event in real practice, ensuring the anomalies are tied to actionable diagnostic contexts. We then apply preprocessing to standardize sampling frequency, normalize scales across sensor channels, and fully de-identify the signals by removing all system identifiers and metadata that could reveal sensitive operational information.

\textbf{Stage 2: Human annotation pipeline.}  
We develop a de-identified annotation pipeline that preserves the realism of paired textual data while protecting privacy. Human experts annotate the selected anomalous windows with concise descriptions of the observed pattern, plausible root causes, and candidate corrective actions. To prevent leakage of proprietary context, annotators are provided only with sanitized time-series segments and high-level machine categories. The resulting annotations capture domain-relevant diagnostic reasoning in natural language while guaranteeing de-identification.

\textbf{Stage 3: Construction of evaluation queries.}  
To enable systematic benchmarking, we cluster the curated time-series into families of similar anomaly types (e.g., belt failure–like patterns vs. thermal runaway patterns). From these families we generate multiple-choice questions that follow a multi-stage structure. Each query involves (i) identifying the anomalous segment, (ii) inferring its root cause, and (iii) suggesting a corrective action. Ground-truth answers are paired with distractors sampled from other clusters, ensuring that solving the task requires both correct recognition and reasoning rather than memorization. 

This multi-stage curation yields \textit{SenTSR-Bench} as a realistic and challenging benchmark, with human-authored annotations grounded in real sensor signals and a design that emphasizes both diagnostic depth and privacy protection.

\subsubsection{Training Dataset Generation}
\label{app:train-data}

Building training data at scale for diagnostic reasoning is especially challenging in the multivariate sensor setting: real-world signals are scarce, and their complexity makes direct augmentation difficult. We therefore propose a two-stage pipeline that leverages vision–language models (VLMs) to bootstrap realistic simulators from a small set of seeds.

\textbf{Stage 1: Iterative code synthesis.}  
We begin with 23 standardized and de-identified multivariate time-series, each containing channels such as vibration (acceleration, velocity) and temperature. Each seed is plotted and presented to a VLM together with high-level context prompts (e.g., “write Python code that simulates similar behavior with interpretable dynamics”). The VLM outputs candidate simulation code, which we execute to generate synthetic traces. If the output contains runtime errors or fails to reproduce core dynamics of the seed (e.g., anomaly shape, periodic structure), we refine the prompt and re-run. This iterative prompt–code–simulate cycle continues until the simulator consistently reproduces the desired behaviors. The outcome is a library of seed-aligned simulators.

\textbf{Stage 2: Diversification and simplification.}  
To scale up diversity, we prompt an LLM to transform each simulator into a stochastic generator. Deterministic heuristics are replaced with latent-state dynamics and randomized parameter draws (e.g., varying noise levels, decay rates, or event frequencies). This produces a family of realistic series rather than exact replicas. We further refactor the simulators into compact, modular forms so they can be easily reused and extended. The diversified generators collectively produce a large corpus of synthetic signals that retain the statistical and structural properties of the real seeds while covering a wider variety of operating conditions.

Finally, we apply the same query-construction pipeline as in evaluation: anomalous segments from synthetic series are paired with diagnostic labels to form QA and MCQ items. This ensures consistency between training and evaluation, while enabling large-scale supervised training from only a handful of seed signals.

\section{Implementation Details}
\label{app:implementation}

\subsection{Implementation Details: Reasoning Model Baselines}
\label{app:baseline-impl}

We evaluate standard reasoning baselines under both zero-shot and few-shot prompting. All models are accessed through an OpenAI-compatible server implemented with \texttt{vLLM}, using HuggingFace checkpoints as backends. Unless otherwise noted, reasoning traces are obtained in a zero-shot setting, while few-shot experiments prepend a small set of curated exemplars. For few-shot prompting, for the \emph{SenTSR-Bench} benchmark, we provide 3 randomly sampled demonstrations in an in-context learning format, inserted as prior user–assistant interactions. Each demonstration contains either the time-series image or JSON text paired with its ground-truth answer. For Qwen3, DeepSeek R1, the encoded time series far exceeds the context length. In such cases, we include only the question and answer template in the demonstrations, omitting the full time-series input.

\paragraph{Encoding time series for LLM input.} 
For image encoding, we render multivariate time series as stacked line plots using \texttt{matplotlib}. Each channel is placed in a vertically aligned subplot with labeled axes and channel identifiers, following best practices for visual clarity. Detailed plotting functions are provided in the released source code. 

For text encoding, we convert each channel into a structured JSON-like format. The following template illustrates the format used to render time-series data into textual tokens for inclusion in prompts:

\begin{lstlisting}[language=json, basicstyle=\ttfamily\small]
{
  "Series 1": [0.25, 0.31, 0.28, ...],
  "Series 2": [1.02, 1.13, 0.95, ...],
  "Series 3": [-0.42, -0.38, -0.41, ...]
}
\end{lstlisting}

This structured form facilitates tokenization and preserves the alignment of values across channels. When column names are available, they are preserved; otherwise, generic names are assigned.

\paragraph{Long-context adaptation.}  
For certain benchmarks such as \emph{TSEvol}, multivariate time-series inputs can exceed \(50\mathrm{k}\) tokens when encoded as text. To accommodate these cases, we apply RoPE scaling to extend the context length of open-source models such as Qwen3 and DeepSeek R1, ensuring that the full series can be processed without truncation. This scaling is necessary for faithfully grounding reasoning in long multivariate signals. 

\paragraph{Infrastructure.}  
All open-source models are hosted on AWS EC2 instances equipped with 8$\times$A100 GPUs, served through \texttt{vLLM}. Closed-source reasoning models are accessed via AWS Bedrock.

\subsection{TSLM Post-training}
\label{app:tslm-training}

All time-series specialists (TSLMs) are initialized from the public \texttt{Qwen-VL-3B-Instruct} checkpoint. Post-training is carried out in two stages: supervised fine-tuning (SFT) and reinforcement learning (RL) with verifiable rewards. For the public benchmarks \emph{TSEvol} and \emph{TS\&Language}, we fine-tune on 3k causal reasoning tasks from the \emph{TSEvol} SFT set, restricting training to causal tasks to test cross-task generalization to inductive, deductive, and MCQ-2 tasks at evaluation. For the \emph{SenTSR-Bench} benchmark, SFT data is constructed from the curated \emph{What happened} and \emph{How happened} stages, leaving the \emph{Suggested fix} stage unseen for out-of-distribution evaluation. SFT training uses a cutoff length of 4{,}096 tokens, per-device batch size of 4 with gradient accumulation of 2 (effective batch size 64 on 8 GPUs), learning rate of \(1\times 10^{-5}\), and cosine decay scheduling with warmup ratio 0.1.

For reinforcement learning, we adopt Group Relative Policy Optimization (GRPO) to elicit analysis-first completions without explicit thinking supervision. For \emph{TSEvol}, RL training again focuses on causal tasks, and for \emph{SenTSR-Bench} we apply it to the \emph{What happened} and \emph{How happened} datasets. RL training is configured with KL divergence coefficient \(\beta=0.001\), group size \(G=8\), maximum sequence length \(L_{\max}=512\), and PPO clipping threshold of 0.1, with an effective batch size of 16 on 8 GPUs, with a learning rate of \(1\times 10^{-6}\). 

\subsection{Practical Implementation of Knowledge Injection.}
The injection workflow is straightforward to implement with standard LLM APIs. For models and servers that support \emph{assistant prefill} (e.g., OpenAI-compatible endpoints), we directly seed the private trace by pre-inserting
\(
[\,\langle\mathrm{think}\rangle,\;\mathbf{r}^{\text{Inj}}_{\le k}\,]
\)
as the assistant’s initial tokens; the general reasoner \(\pi^{G}\) then continues generation conditioned on this prefix. For providers that do not expose editable thinking buffers (in some closed-source reasoning models), we use an {instructional proxy}: wrap \(\ \mathbf{r}^{\text{Inj}}_{\le k}\) inside the models’s recommended “thinking template” tags in the user/system message (e.g., a documented \texttt{<thinking>}…\texttt{</thinking>} block) and instruct the model to begin its thinking process with the instructed template. In practice this proxy reliably steers the internal reasoning trace and reproduces the effect of in-chain injection. 

\subsection{Prompt Design for Injection Strategies}

We provide the prompt templates used in experiments for evaluating different \emph{knowledge injection positions}. These prompts are designed for a strong instruction-following TSLM (\texttt{ChatTS-14B}) paired with general reasoning LLMs (GRLMs). The goal is to examine how injecting time-series knowledge at different points in the reasoning process, including \emph{early}, \emph{intermediate}, and \emph{late} injection, affects overall reasoning performance.

\paragraph{Early Injection.}
In the early injection setup, the TSLM first produces structured, quantitative observations from the time series, which are inserted at the start of the GRLM’s reasoning process to guide the subsequent chain of thought.

\begin{promptbox}{TSLM (Observation Generation)}
You are analyzing a time series to extract key quantitative observations that help answer the question.  
Provide detailed, objective numerical observations by following these guidelines:  
1. Make numbered, precise observations about the quantitative aspects of the time series.  
2. Be specific about values, positions, and magnitudes when describing features.  
3. Begin each observation with "Observation 1:", "Observation 2:", etc.  

Start your response with:  
\textit{"To answer this question, I need to carefully analyze the time series. Here are my observations: Observation 1... Observation 2..."}  
\end{promptbox}

\begin{promptbox}{GRLM (Reasoning with Early Injection)}
<think> [TSLM Observations]  
Wait, let me summarize and reflect on the previous observations from the time series,  
and then continue my reasoning process to derive the final answer...
\end{promptbox}

\paragraph{Intermediate Injection.}
In this setting, the GRLM begins reasoning but calls for the TSLM’s input when encountering uncertainty (identified as a low-confidence step). The TSLM then provides clarifications, which are integrated back into the GRLM’s ongoing reasoning.

\begin{promptbox}{TSLM (Intermediate Feedback)}
You are assisting with a time series reasoning process.  
Here is the question and the current partial reasoning:  
[Partial Thought / Low-Confidence Segment]  
Please analyze whether this reasoning point is correct based on the time series, why or why not,  
and how it relates to answering the question. Be specific about numerical values and time positions.  
\end{promptbox}

\begin{promptbox}{GRLM (Reasoning with Intermediate Injection)}
<think>  
[Partial Reasoning]  
I am uncertain about this point: [Low-Confidence Segment].  
Let me reconsider it with the following clarification from the time-series model:  
"[TSLM Feedback]" ...
\end{promptbox}

\paragraph{Late Injection.}
Under late injection, the GRLM first completes its reasoning trace.  
The TSLM then reviews the reasoning for factual consistency with the time series, and the GRLM revises its conclusion accordingly.

\begin{promptbox}{TSLM (Late Review)}
You are reviewing a completed reasoning process to check whether its quantitative claims about the time series are correct.  
For each observation, discuss whether it is correct or incorrect.  
If incorrect, provide the accurate interpretation of what the data shows.  
Example:  
"For Observation 1, after review, it is incorrect. In fact... For Observation 2..."  
\end{promptbox}

\begin{promptbox}{GRLM (Revision with Late Injection)}
<think>  
[Original Reasoning Trace]  
Wait, let me reexamine my previous reasoning based on the review below:  
[TSLM Review]  
...
\end{promptbox}

\paragraph{Additional Adaptations.}

For the \textbf{RL-honed TSLM}, we apply \emph{early injection} by directly using the model’s self-generated reasoning trace from R1-style GRPO training as the injected knowledge, without explicit prompting; for \textbf{closed-source models} (e.g., \texttt{Claude-3.7}), injected content is wrapped in \texttt{<thinking>...</thinking>} delimiters, followed by an instruction such as “continue the thinking process above.”; for the \textbf{prompting-based baseline}, the same TSLM-generated content is provided externally as additional context:  

\begin{promptbox}{Prompting Baseline}
Here is an analysis from a time-series model that is good at time-series analysis  
but may be limited in general reasoning: [TSLM Observations].  
\end{promptbox}

\section{Additional Case Study}
\label{app:case_study}
\label{app:inject_over_prompt}

\begin{figure}
    \centering
    \includegraphics[width=1.0\linewidth]{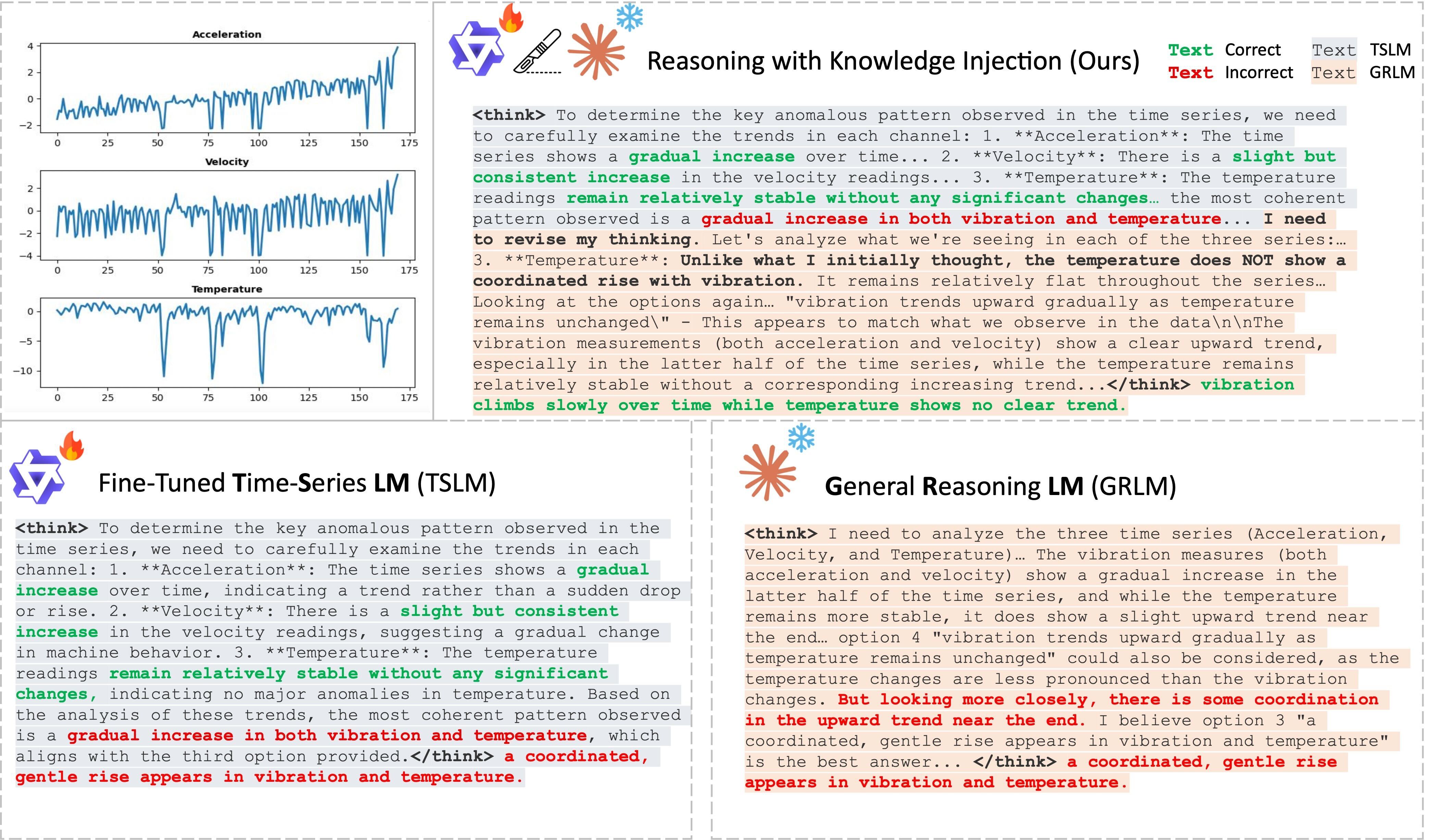}
    \caption{Case study on \textit{knowledge injection} versus standalone baselines. The \textbf{TSLM} correctly detects rising vibration and stable temperature but hallucinates a joint increase in both, yielding an incorrect diagnosis. The \textbf{GRLM} similarly misreads the series, assuming a late temperature rise. Our method leverages the TSLM’s accurate signal interpretation while correcting its reasoning flaw, producing the correct final diagnosis.}
    \label{fig:case_study_injection}
\end{figure}

\begin{figure}
    \centering
    \includegraphics[width=1.0\linewidth]{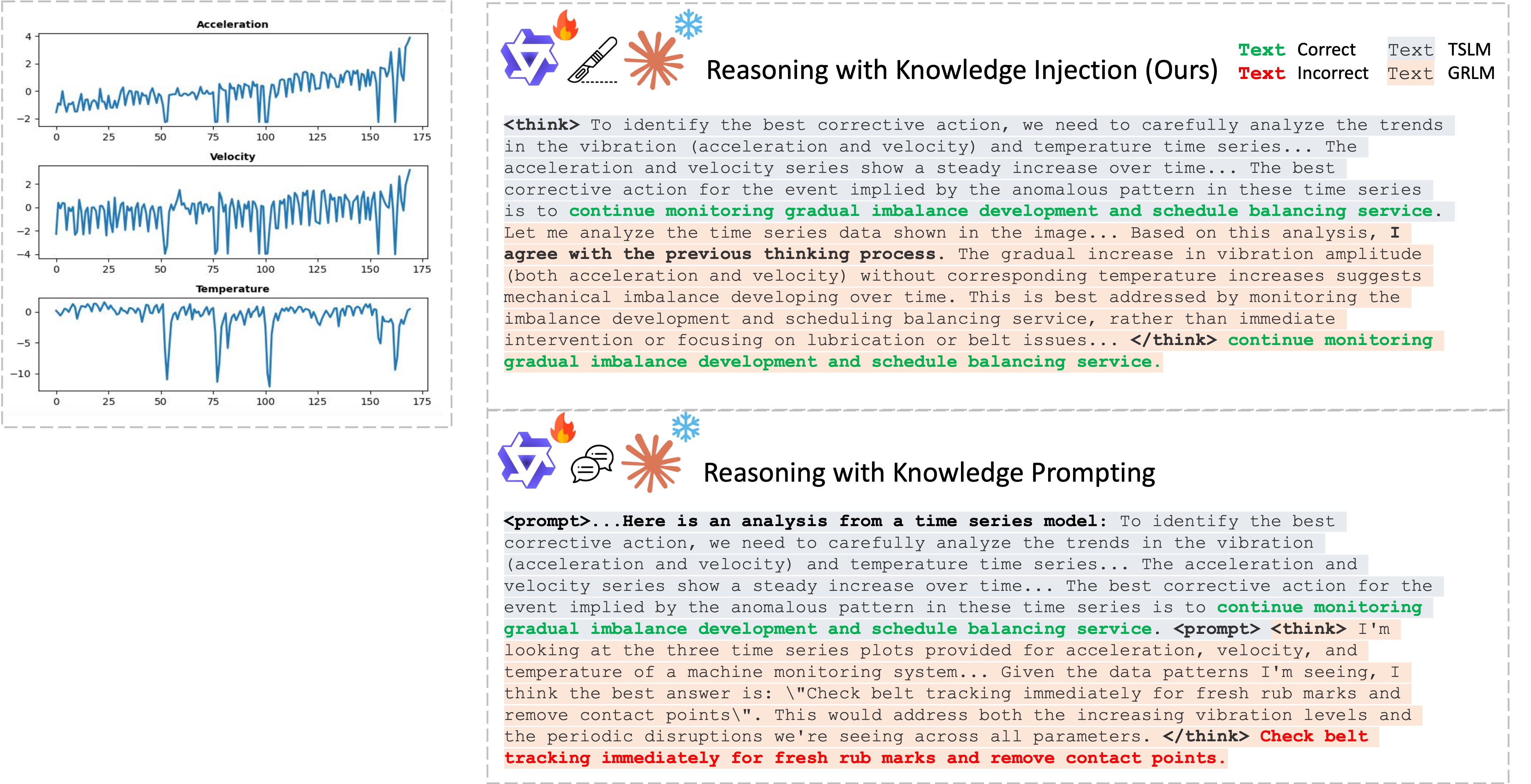}
    \caption{Case study on knowledge prompting versus knowledge injection. In the prompting-based approach, the \textbf{GRLM} reasons largely in isolation, referencing the \textbf{TSLM}’s analysis only at the end for validation, leading to partial use of domain knowledge. In contrast, the injection-based approach integrates the \textbf{TSLM}’s discussion directly into the reasoning flow, enabling joint exploration and narrowing of hypotheses, and resulting in a correct, well-grounded diagnosis.}
    \label{fig:case_study_prompting}
\end{figure}

We present two qualitative case studies that illustrate the benefits of our knowledge injection framework. Figure~\ref{fig:case_study_injection} compares the standalone TSLM, the standalone GRLM, and our injection-based method on a diagnostic reasoning example. The TSLM correctly detects rising vibration and stable temperature but hallucinates a joint increase in both signals, yielding an incorrect diagnosis. The GRLM similarly misreads the series, assuming a late temperature rise. By injecting the TSLM’s accurate signal-level observations into the GRLM’s reasoning trace, our method corrects the reasoning flaw while preserving domain-grounded pattern recognition, producing the correct final diagnosis. Figure~\ref{fig:case_study_prompting} further contrasts knowledge \emph{prompting} with knowledge \emph{injection}. When the TSLM analysis is provided as an external prompt, the GRLM reasons largely in isolation, leading to insufficient use of domain knowledge. In contrast, the injection-based approach integrates the TSLM’s analysis directly into the reasoning flow, enabling joint exploration and progressive narrowing of hypotheses, and resulting in a correct diagnosis.

\end{document}